\documentclass[acmtog,authorversion,nonacm]{acmart}
\usepackage{subfigure}
\usepackage{siunitx}
\usepackage{cleveref}
\usepackage[utf8]{inputenc}
\usepackage{framed}
\usepackage{listings}
\usepackage{tabularx}
\usepackage{multirow}
\usepackage{xcolor} \usepackage{float} \usepackage{caption}

\usepackage{tikz}

\definecolor{beige}{RGB}{230, 160, 118}
\definecolor{orange}{RGB}{230, 60, 0}

\AtBeginDocument{  \providecommand\BibTeX{{    \normalfont B\kern-0.5em{\scshape i\kern-0.25em b}\kern-0.8em\TeX}}}

\acmSubmissionID{2366}

\begin{document}
\setcopyright{none}
\acmJournal{TOG}
\acmYear{2026}
\acmVolume{45}
\acmNumber{6}
\acmArticle{241}
\acmMonth{12}
\acmDOI{10.1145/3842576}

\title{ProAct: Harnessing Streaming Motion Generation and Agentic Reasoning for Real-Time Embodied Social Interaction}

\author{Zeyi Zhang}
\authornote{equal contribution}
\email{illusence1@gmail.com}

\author{Zixi Kang}
\authornotemark[1]
\email{forever.kzx0713@stu.pku.edu.cn}

\author{Ruijie Zhao}
\authornotemark[1]
\email{zhaoruijie@stu.pku.edu.cn}
\affiliation{    \institution{Peking University}
    \country{China}
}

\author{Yusen Feng}
\email{ysfeng@stu.pku.edu.cn}

\author{Biao Jiang}
\email{bjiang25@stu.pku.edu.cn}

\author{Hanyu Ji}
\email{jih484287@gmail.com}

\author{Libin Liu}
\authornote{corresponding author}
\email{libin.liu@pku.edu.cn}
\orcid{0000-0003-2280-6817}
\affiliation{  \institution{Peking University}
  \country{China}
}

\begin{abstract}

Real-time embodied social interaction places two equally demanding requirements on an agent: continuously generating fluent multimodal interaction behavior, and proactively reasoning over accumulated dialogue and visual context to decide when to take initiative. These requirements must both be satisfied under a strict latency budget, making them difficult to meet simultaneously. We present \emph{ProAct}, a dual-system framework that manages these time-critical requirements by integrating a low-latency \emph{Behavioral System} for streaming multimodal interaction with a slower \emph{Cognitive System} that performs long-horizon social reasoning and produces high-level proactive intentions. The Cognitive System incorporates an efficient memory mechanism and a user-motivation prediction module to reason over accumulated dialogue and visual context and determine when proactive intervention is appropriate. The Behavioral System further includes an intention-conditioned streaming flow-matching motion generator with a disentangled ControlNet branch, which translates deliberative intentions into continuous non-verbal behavior without disrupting interaction fluency.
We deploy ProAct on a physical humanoid robot and validate the framework through comprehensive experiments, including real-world user studies, motion-generation benchmarks, and evaluation on \emph{ProActBench}, a new, targeted benchmark for evaluating proactive trigger detection and restraint in embodied interaction.
Code and ProActBench will be released at \url{https://proactrobot.github.io/}.

\end{abstract}



\begin{teaserfigure}
  \centering
  \includegraphics[width=\textwidth]{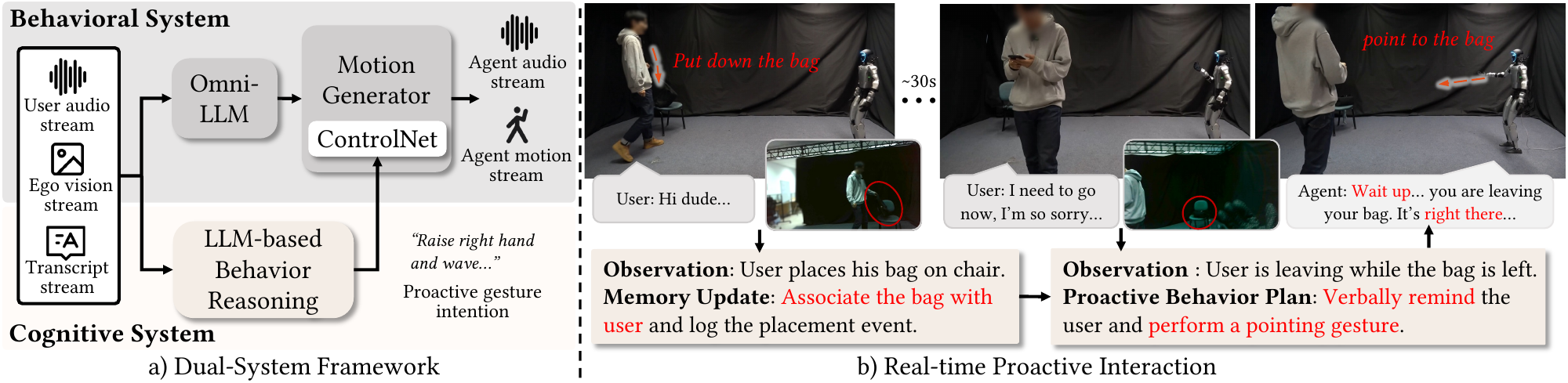}
  \caption{ProAct enables proactive embodied social interaction. (a) Our dual-system framework integrates a fast Behavioral System for streaming multimodal generation with a Cognitive System for context-driven reasoning. (b) In a Real-time Interaction scenario, the agent tracks long-term context and autonomously initiates a synchronized verbal and gestural reminder when the user attempts to leave, demonstrating proactivity beyond simple reactivity.}
  \Description{Overview of the ProAct dual-system framework and a qualitative example where the embodied agent proactively reminds a user about a forgotten item during real-time interaction.}
  \label{fig:teaser}
\end{teaserfigure}

\maketitle

\section{Introduction}
\label{sec:intro}

Natural social interaction is inherently bidirectional. When conversing with others, humans do not merely respond to what is said. They anticipate needs, take initiative, and act before being asked~\cite{bandura2001social}: a host notices empty glasses and offers a refill, and a friend senses hesitation and provides encouragement without prompting. This \emph{proactivity}---the capacity to originate actions from internal goals rather than passively respond---distinguishes genuine social partners from mere responders.

For embodied agents operating in the real world, such proactivity must be achieved under strict timing constraints: the agent must reason over accumulated dialogue and visual context to decide when and how to intervene, while continuously producing fluent multimodal behavior that preserves real-time embodied presence throughout the interaction~\cite{vandenbroek2024proactive}.  Crucially, both processes must be completed within a tight latency budget: delayed reasoning can miss the appropriate moment for intervention, while delayed motion generation can break conversational timing and disrupt embodied presence.

Turning this notion into a deployable system runs into two interlocking technical challenges that must be solved jointly: a cognitive challenge of deciding \emph{when and what} the agent should proactively do, and a behavioral challenge of realizing \emph{how} that decision becomes a continuous, real-time multimodal motion stream.

\emph{At the cognitive layer}, proactive decision-making requires reasoning over accumulated dialogue history and visual scene context at a human-plausible social timescale.
This reasoning is fundamentally slower than the low-latency loop that real-time interaction demands, and a monolithic model faces a bidirectional trade-off: prioritizing immediate response yields reflexive behaviors that lack long-term social coherence, while deliberative models introduce latency that breaks conversational fluency.
Our goal is therefore not to eliminate deliberation latency, but to prevent it from blocking the real-time behavioral loop. Dual-process theories in cognitive science~\cite{watson2011d} and recent dual-system architectures in autonomous driving~\cite{tian2024drivevlm} and robotics~\cite{figure_helix_2025} suggest that separating immediate reactive execution from slower reasoning is a viable path to making both regimes coexist.

\emph{At the behavioral layer}, the agent must continuously synthesize multimodal behavior in real time. Existing streaming gesture generators meet this real-time constraint but are audio-driven only, while text- or style-conditioned generators~\cite{yang2024Freetalker,chen2024syntalker,SocialAgent2025SIGA} accept semantic guidance only offline, outside real-time stream. Proactive embodiment requires both at once: when the cognitive layer decides the agent should point to a forgotten bag, the new intention must enter the audio-driven motion stream mid-flight, without desynchronizing it. The behavioral layer therefore needs a streaming, real-time motion generator that admits asynchronous, dual-channel conditioning within a single continuous trajectory.

We present \emph{ProAct}, a dual-system framework that addresses both bottlenecks jointly.
A fast \emph{Behavioral System} handles the real-time interaction loop. It integrates a streaming omni-modal LLM for verbal response with a streaming motion generator for continuous non-verbal behavior. The motion generator is built on flow matching with a disentangled ControlNet branch, so it supports \emph{dual conditioning}: audio rhythms drive reactive co-speech gestures, while semantic intentions from the Cognitive System can be asynchronously injected into the same stream. An overlap-and-cache scheme (Sec.~\ref{sec:mogen}) keeps the stream continuous across chunk boundaries even while conditioning switches, so reactive and proactive gestures are mixed within one continuous motion trajectory rather than played back as separate clips.
A slower \emph{Cognitive System} runs in parallel as an LLM-driven agentic framework. It maintains a bounded memory bank, scores the current context along five motivation dimensions, and emits proactive intentions when a dimension clears a threshold---without blocking the behavioral loop on its slower inference.

We deploy the full system on a physical humanoid and validate it from the complete system down to its two core modules. First, real-world user studies evaluate the complete dual-system interaction and show that participants and observers rate ProAct higher than reactive baselines on \emph{Active Agency}, perceived presence, and interaction comfort. We then analyze the fast Behavioral System through motion-generation experiments, where ProAct improves over state-of-the-art gesture models on FGD, BeatAlign, and diversity. Finally, to examine the slow Cognitive System more directly, we evaluate ProAct on \emph{ProActBench} (Sec.~\ref{sec:proactbench_eval}), a new, targeted benchmark of human-annotated scripted scenarios for evaluating when an embodied agent should proactively intervene or deliberately refrain. Ablations on this benchmark support the effectiveness of our structured cognitive harness for proactive decision-making and interaction quality.

Our contributions are as follows:
\begin{itemize}
  \item A dual-system architecture for embodied social agents that lets slower context-driven proactive reasoning run alongside real-time reactive multimodal generation without blocking the behavioral loop.
  \item An intention-conditioned streaming motion generator based on flow matching with a disentangled ControlNet branch, which supports dual conditioning on audio rhythm and semantic intent and lets reactive and proactive gestures be generated within a single continuous motion stream rather than as separately scheduled clips.
  \item A complete physical-humanoid deployment, validated through comprehensive experiments that evaluate both motion quality and interactive effectiveness.
\end{itemize}

\section{Related Work}
\label{sec:related}

\subsection{Embodied Social Interaction Systems}

Interactive agents vary in their degree of embodiment and interaction modality. \emph{Text- or speech-based conversational agents}~\cite{shao2023character, characterai, kyutai2024moshi} enable rich linguistic exchange through modern LLMs, but lack embodiment for non-verbal expression and environmental interaction.
\emph{Virtual embodied agents} address this limitation by generating synchronized speech and nonverbal behavior for digital characters. These systems either employ LLM-cascaded models where speech conditions downstream nonverbal behavior generation~\cite{kim2024bodyagent, dlp3d, ai_character_platform,cai2025interactiveintelligencedigitalhumans,cai2023digitallifeprojectautonomous}, or encode multimodal behavior jointly within the LLM backbone~\cite{ao2024body, jiang2025solami,zhang2025vibesconversationalagentbehaviorallyintelligent}. While achieving impressive realism, they remain confined to virtual environments. \emph{Physically embodied robots} extend interaction into the real world. Motion-based interaction frameworks~\cite{chen2025symbiosimhumanintheloopsimulationplatform, ji2025immersivehumanxinteractionrealtime} detect human intent from visual cues and generate responsive physical actions, enabling real-time collaboration. Commercial humanoid companion robots~\cite{EngineeredArts_Ameca, HansonRobotics_Sophia, kang2024nadinellmdrivenintelligentsocial} emphasize facial animation and scripted speech, but typically lack expressive full-body motion generation.

While these systems achieve real-time responsiveness, they remain signal-driven: motion generation is directly coupled to immediate sensory input like user motion and speech audio without incorporating high-level reasoning. Proactive social behavior, in contrast, requires reasoning over accumulated conversational context to infer when and how to initiate unsolicited but contextually appropriate actions.
Recent work studies proactive assistance with LLM agents and streaming egocentric video~\cite{lu2025proactive,zhang-etal-2025-proactive}, but does not address its realization through continuous whole-body behavior.

\subsection{Non-verbal Behavior Generation}

Non-verbal behavior generation has been extensively studied, particularly in the context of co-speech gestures~\cite{Nyatsanga_2023}, where diverse generative architectures—ranging from statistical and neural models to Transformers and diffusion approaches—have enabled increasingly natural gesture synthesis~\cite{yoon2020trimodalgesture, alexanderson2023listendenoiseaction, ao2022rhythmic, ao2023gesturediffuclip, yi2022talkshow, liu2023emage, ng2024audio2photoreal, 10.1145/3721238.3730611, zhang2025semtalk,liu2025semges,siggesture2024,ghorbani2022zeroeggs,liu2022beat, cheng2025hop, yinpyramotion, Yang_2025_ICCV, Mughal_2025_CVPR}.
Building on these foundations, recent work has expanded toward dyadic or partner-aware gesture generation~\cite{mughal2024convofusion, qi2025co3gesturecoherentconcurrentcospeech, sun2024beyondtalking, SocialAgent2025SIGA,echo_siga25,agrawal2025seamlessinteractiondyadicaudiovisual,mclean2025embody3dlargescalemultimodal}, multimodal control synthesis guided by text instructions or semantic signals beyond raw audio~\cite{zhang2024semantic, yang2024Freetalker, chen2024syntalker, chen2024body_of_language}, and real-time and arbitrary-length generation, with efficient network architectures~\cite{liu2025gesturelsmlatentshortcutbased}, and streaming generation strategy through autoregressive formulations or rolling diffusion~\cite{zhen2025tellerrealtimestreamingaudiodriven,Xiao_2025_ICCV, cai2025flooddiffusiontailoreddiffusionforcing,tseng2022edge} to achieve frame-by-frame synthesis with low latency.
Our work integrates these capabilities through a streaming flow-matching architecture with ControlNet-based dynamic conditioning, supporting real-time gesture synthesis modulated by asynchronously injected control signals.

\subsection{Dual System for Real-time Interaction}

Dual-system architectures, inspired by the distinction between fast reactive \emph{System-1} and slower proactive \emph{System-2} reasoning~\cite{watson2011d}, reconcile the tension between low-latency responsiveness and contextually grounded decision-making.
A single monolithic model often struggles to satisfy both: fast models lack semantic depth, while deliberative models are too slow for interactive control.
To address this, conversational agents such as the Talker–Reasoner framework~\cite{christakopoulou2024agents} separate immediate dialogue generation from slower multi-step reasoning. In autonomous driving, DriveVLM-Dual~\cite{tian2024drivevlm} integrates high-frequency visuomotor control with a slower VLM-based module to support semantic scene understanding and long-horizon planning. In robot control, Helix~\cite{cui2025openhelix} similarly pairs a high-rate reactive controller with a slower VLM that provides task-level objectives and semantic grounding.
We apply this design to embodied social interaction: our Behavioral System (System-1) generates streaming multimodal responses, while our Cognitive System (System-2) injects proactive intentions through social reasoning.

\section{Overview}
\label{sec:overview}

As illustrated in Figure \ref{fig:teaser} a), \emph{ProAct} adopts a dual-system architecture that couples a fast, reactive {Behavioral System} with a
planning-oriented, proactive {Cognitive System}. The {Behavioral System} operates as the real-time interaction loop: it continuously processes user audio and visual inputs and produces low-latency multimodal responses.
It integrates a streaming omni-modal LLM for generating verbal responses with a streaming
motion generator for continuous synthesis of non-verbal behavior.
While speech is generated in turn-based segments after the user finishes speaking, the motion generator continuously produces non-verbal behavior from both user and agent audio streams, ensuring uninterrupted motion even during listening phases.

Running in parallel, the {Cognitive System} performs social reasoning over accumulated context via an agentic framework with episodic memory.
This framework observes the ongoing dialogue and visual scene, and infers when and what proactive actions should be initiated.
When triggered, it generates intention signals, which are injected into the {Behavioral System} to modulate both verbal and non-verbal responses.
This allows the robot to combine fast reactive behaviors with deliberate, context-driven proactive actions.

\section{Behavioral System: Streaming Interaction}
\label{subsec:system1}

The Behavioral System maintains the real-time interaction loop, producing streaming verbal and non-verbal responses through a cascaded architecture that connects a streaming omni-modal LLM with a streaming motion generator. The two channels operate asynchronously: verbal responses follow a turn-based pattern, while motion generation runs continuously to maintain embodied presence throughout the interaction.
We deploy the generated motion on a physical robot through a hierarchical whole-body control policy. The generated upper-body joint angles are streamed to the robot for real-time tracking, while lower-body balance is maintained by the built-in balance and locomotion controllers.

\begin{figure*}[t]
    \centering
    \includegraphics[width=0.80\textwidth]{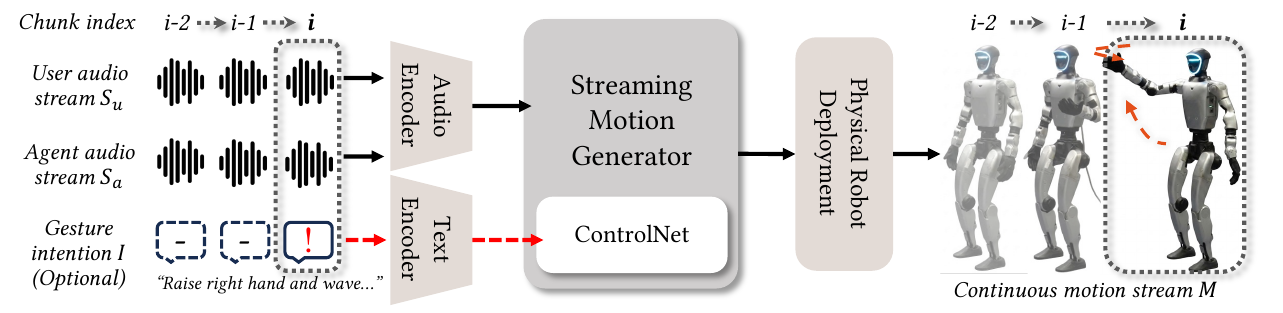}
    \caption{\textbf{Real-time streaming motion generation in the fast Behavioral System.} At each chunk index $i$ of Eq.~\ref{eq:model}, the user and agent audio streams $S_{\mathrm{u}}^{i}, S_{\mathrm{a}}^{i}$ condition the Streaming Motion Generator to produce the next motion chunk $M^{i}$, with the last $\ell$ frames of $M^{i-1}$ reused for temporal continuity. When a gesture intention $I^{i}$ is provided, the disentangled ControlNet branch injects its semantics into the ongoing chunk so the intended gesture appears without breaking the continuous motion stream.}
    \label{fig:streaming_motion}
\end{figure*}

\subsection{Streaming Verbal Response Synthesis}

The omni-modal LLM processes user speech and visual observations from the robot's egocentric camera to generate audio responses.
We use GPT Realtime model~\cite{openai2025gptrealtime}, which operates in turn-based mode with streaming input and output. User audio is continuously streamed chunk-by-chunk for low-latency processing. Upon detecting turn completion via voice activity detection, the model begins generating agent audio in a streaming fashion, producing chunks progressively for immediate playback. The model supports mid-utterance interruption if the user speaks during the agent's response, ensuring interaction fluency.

\subsection{Controllable Streaming Motion Synthesis}
\label{sec:mogen}
We represent motion as per-frame joint rotations in a unified feature space (see Sec.~\ref{subsubsec:data}).
In contrast to the turn-based verbal channel, motion generation operates continuously throughout the interaction to maintain embodied presence.
The motion generator must meet two requirements. First, \emph{real-time generation}: the system must synthesize motion faster than its playback duration, ensuring continuous, uninterrupted motion streaming after the initial response. Specifically, for a motion sequence of duration $T_{\mathrm{motion}}$, the computation time for generation $T_{\mathrm{gen}}$ must satisfy $T_{\mathrm{gen}} < T_{\mathrm{motion}}$. Second, \emph{controllable generation}: the system must follow high-level semantic motion instructions from Cognitive System to enable proactive behavior intentions. To address these challenges, we propose a controllable streaming motion generator based on flow matching.

Formally, we cast motion generation as a streaming problem that operates chunk-by-chunk generation with $n$ frames for each chunk. At the $i$-th generation step, the generator $\mathcal{G}$ takes synchronized dyadic speech chunks ${S}_{\mathrm{u}}^{i},\, {S}_{\mathrm{a}}^{i}$ from both user and agent, an optional intention instruction ${I}^{i}$, and the last $\ell$ frames of the previous chunk ${M}^{i-1}$ (denoted as ${M}^{i-1}_{-\ell:}$) to ensure temporal continuity:
\begin{equation}
{M}^{i}
= \mathcal{G}\!\left(
{S}_{\mathrm{u}}^{i},
{S}_{\mathrm{a}}^{i},
{I}^{i},
{M}^{i-1}_{-\ell:}
\right).
\label{eq:model}
\end{equation}
Figure~\ref{fig:streaming_motion} illustrates this chunk-by-chunk formulation and the injection of gesture intentions without breaking the motion stream.

For a detailed illustration of the model architecture and specific training and inference strategies, please refer to Appendix~\ref{appendix:model_arch}.

\subsubsection{Flow Matching for Real-time Motion Synthesis}
\label{sec:flow_matching}

To model the motion distribution ${M}$ under real-time constraints, we adopt Conditional Flow Matching~\cite{lipman2023flowmatchinggenerativemodeling}, which transforms noise into data through a learned continuous-time flow.
Specifically, flow matching learns a time-dependent velocity field $v_\tau({M}_\tau, \mathcal{C})$ that transports samples from Gaussian noise ${M}_0 \sim \mathcal{N}({0}, {I})$ at $\tau=0$ to the data distribution ${M}_1 \sim q_{\mathrm{data}}$ at $\tau=1$, where the conditioning variable $\mathcal{C}$ comprises the audio chunks. The generation follows an ordinary differential equation (ODE):
\begin{equation}
\frac{d{M}_\tau}{d\tau} = v_\tau({M}_\tau, \mathcal{C}),
\quad \tau \in [0,1].
\label{eq:ode}
\end{equation}

We use the optimal-transport path that linearly interpolates between noise and data: ${M}_\tau = \tau {M}_1 + (1-\tau){M}_0$. We train a neural velocity estimator $v_\theta$ to match the path velocity ${M}_1 - {M}_0$ by minimizing:
\begin{equation}
\mathcal{L}_{\mathrm{CFM}}(\theta) = \mathbb{E}_{\tau, {M}_1, {M}_0}\left[\|v_\theta(\tau, {M}_\tau, \mathcal{C}) - ({M}_1 - {M}_0)\|^2\right],
\label{eq:cfm_loss}
\end{equation}
where $\tau \sim \mathcal{U}[0,1]$. During inference, we sample a noise point $M_0$ and iteratively solve the ODE to generate $M_1$ using Euler integration.
This straight-line path simplifies ODE integration compared to diffusion models~\cite{DDPM}, enabling high-quality generation with fewer solver steps, thus achieving faster generation without compromising motion quality.

We instantiate the velocity estimator $v_\theta$ as $\mathcal{G}$, a transformer-based backbone built on DiT blocks~\cite{DiTiccv23}. Unlike prior methods that generate gestures from a single speaker's audio, $\mathcal{G}$ explicitly models dyadic interaction through two synchronized audio streams ${S}_{\mathrm{a}}$ and ${S}_{\mathrm{u}}$. This enables the agent to produce both speaker-side expressive gestures and listener-side attentive behaviors such as subtle body sway and attentive idling during partner speech. The model processes noisy motion ${M}_\tau$ through stacked DiT blocks, each containing self-attention and feed-forward layers with residual connections. The two audio streams are concatenated channel-wise, with their positions in the feature dimension implicitly distinguishing self versus partner audio. The audio features and the time embeddings of the flow are concatenated and injected via AdaLN-Zero~\cite{DiTiccv23} to modulate the generation process, ensuring temporal consistency across the flow trajectory.

To ensure temporal continuity across streaming windows, we use an overlap-and-cache scheme enabled by the tileable property of flow matching~\cite{tseng2022edge}. We cache the last $\ell$ frames from the previous window, prepend them to the next window as an $\ell$-frame overlap, and clamp this overlap at every solver step by overwriting it with the cached frames. This naturally preserves recent motion history during flow evolution, eliminating boundary discontinuities. Combined with flow matching's low-step sampling, our system generates continuous motion sequences faster than the playback duration, satisfying the real-time criterion.

\begin{figure*}[tp]
    \centering
    \includegraphics[width=\linewidth]{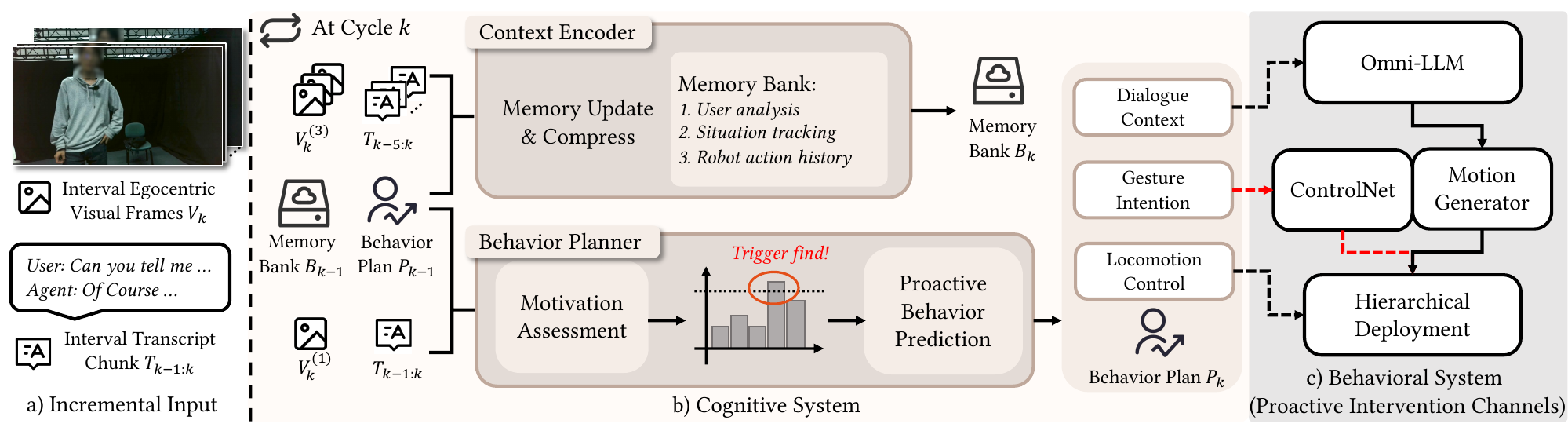}
    \caption{\textbf{Architecture of the Cognitive System.} The system operates in continuous cycles, with each cycle consisting of three stages: (a) collecting incremental multimodal inputs; (b) parallel execution of Context Encoder (memory compression) and Behavior Planner (proactive behavior prediction); (c) injecting behavior plans into the Behavioral System via different channels.}

    \label{fig:method}
\end{figure*}

\subsubsection{Semantic Intention Control via Disentangled ControlNet}
\label{sec:controlnet}

To empower the agent with proactive non-verbal capabilities, the motion generator must support dual conditioning: responding to audio rhythms while adhering to high-level semantic intentions from the Cognitive System. These intentions are formulated as natural language descriptions, requiring the generator to function as a text-to-motion synthesis component operating in parallel with the audio-driven process.
Training a unified model often requires paired audio-motion-text data, where speech audio modulates the rhythm and fine-grained semantics of the motion, while text descriptions provide high-level semantic guidance. However, such datasets are largely absent, posing a critical challenge.
Prior work~\cite{yang2024Freetalker, chen2024syntalker} addresses this by jointly training on separate audio-motion and text-motion datasets within a single shared backbone.
However, our preliminary experiments reveal that this approach suffers from an alignment issue: the model struggles to disentangle rhythmic synchronization from semantic understanding, leading to compromised performance (see ablation in Sec.~\ref{subsubsec:training}).

Inspired by the recent success of ControlNet~\cite{zhang2023addingconditionalcontroltexttoimage} in motion synthesis~\cite{motionlcm}, we introduce a disentangled ControlNet architecture that decouples semantic control from rhythmic generation.
Our system comprises a frozen, pre-trained audio-driven base generator and a parallel trainable control branch initialized as a structural copy. Semantic intention signals are encoded and injected into the control branch, preserving the original audio-driven priors while enabling explicit semantic guidance.

During inference, the system switches between audio-driven and intention-driven behaviors: when no intention is provided, the zero-initialized connections keep the control branch close to zero, so generation falls back to the audio-driven base model.
Because the ControlNet branch shares the same lightweight backbone as the base flow-matching generator, it introduces only a small constant overhead. As a result, our system maintains real-time generation speed while enabling flexible intention control in a unified model.

\section{Cognitive System: Social Context Reasoning}
\label{subsec:system2}

While the Behavioral System ensures fluent reactive interaction, truly social agents must also exhibit proactive behaviors when contextually appropriate. The Cognitive System addresses this need through an LLM-based agentic framework that performs deliberative social reasoning in parallel with real-time interaction.
While operating at a slower pace than the Behavioral System to allow reasoning over accumulated context, it faces a competing constraint: excessive latency risks missing timely intervention opportunities.
Moreover, extended interactions generate increasingly long dialogue and reasoning histories, leading to progressively increasing inference time.

To address the inference time control problem, we decompose the reasoning process into two specialized components: a \emph{Context Encoder} that compresses the accumulating interaction history into a bounded memory bank, and a \emph{Behavior Planner} that performs motivation assessment and action planning.
As illustrated in Figure \ref{fig:method}, within each reasoning cycle, both components operate in parallel on outputs from the previous cycle. By limiting the context length through compression, we ensure that each cycle completes within a consistent time budget $t_c$ regardless of conversation length.
The Cognitive System executes these reasoning cycles continuously in a streaming fashion. Each cycle is triggered immediately upon completion of the previous one, ensuring temporal coherence while maintaining a steady reasoning cadence for timely proactive responses.
The detailed prompts are provided in Appendix~\ref{appendix_Prompts}.

\subsection{Context Encoder}

The Context Encoder maintains a compressed memory bank to prevent unbounded context growth.
At the start of cycle $k$, the encoder receives three incremental inputs: dialogue transcripts ${T}_{k-5:k}$ accumulated over the five most recent inter-cycle intervals for sufficient temporal context, three visual frames ${V}_k^{(3)}$ uniformly sampled since the last cycle to capture scene dynamics, and the previous behavior plan ${P}_{k-1}$ from the Behavior Planner.
The encoder analyzes these inputs to extract salient events and observations, then integrates them into the existing memory bank ${B}_{k-1}$ through incremental summarization, producing an updated compressed ${B}_k$ of a bounded size: ${B}_k = \text{ContextEncoder}({B}_{k-1}; {T}_{k-5:k}, {V}_k^{(3)}, {P}_{k-1})$.

The memory bank ${B}_k$ maintains a structured representation across three hierarchical levels. First, inspired by Theory of Mind reasoning~\cite{Premack_Woodruff_1978_tom}, \textit{user analysis} tracks the user's observable appearance and inferred mental state, capturing their beliefs, desires, and behavioral intentions. Second, \textit{situation tracking} records significant events, observable environmental changes, and unresolved tasks as they emerge during interaction. Third, \textit{robot action history} summarizes previously executed proactive behaviors to prevent redundant interventions. To control memory bank size, each component within these levels is constrained to a single-sentence summary. This compression-based approach trades perfect recall for consistent inference speed.

\subsection{Behavior Planner}

The Behavior Planner determines when and how to initiate proactive actions based on the compressed memory and current observations. At each cycle $k$, the planner operates on the most recent information: it receives the dialogue transcript ${T}_{k-1:k}$ accumulated since the previous cycle, the most recent visual frame ${V}_k^{(1)}$, and the memory bank ${B}_{k-1}$ and behavior plan ${P}_{k-1}$ from the previous cycle. This design enables responsive decision-making based on immediate context while leveraging compressed long-term memory for contextual grounding. We employ an asymmetric input window to balance performance: this allocates computational budget to accurate history summarization while minimizing Planner latency for timely interventions. The planner outputs a plan ${P}_k$ specifying a set of proactive actions: ${P}_k = \text{BehaviorPlanner}({T}_{k-1:k}, {V}_k^{(1)}, {B}_{k-1}, {P}_{k-1})$.

\subsubsection{Motivation Assessment}

Determining when to intervene is non-trivial: acting too frequently disrupts conversational flow, while inaction misses opportunities for engagement. Inspired by prior work on proactive behavior prediction~\cite{liu2025proactiveconversationalagentsinner}, we design a five-dimensional motivation assessment module where each dimension captures a distinct contextual trigger:
\begin{itemize}
    \item \textit{Visual Scene Changes}: Monitoring dynamic changes in the environment that may require attention.
    \item \textit{User Intent Signals}: Detecting explicit or implicit cues in user behavior suggesting a need for assistance.
    \item \textit{Conversation State}: Analyzing dialogue flow to identify appropriate moments for intervention.
    \item \textit{Social Protocol Requirements}: Adhering to social norms and rituals in daily interaction.
    \item \textit{Emotional Response Needs}: Recognizing the user's emotional state and responding appropriately.
\end{itemize}

The Motivation Assessment module scores each dimension from 1 to 5 based on contextual salience. Let $s_k^{(d)}\in\{1,\dots,5\}$ denote the score of dimension $d$ at cycle $k$, and define the cycle-level motivation score as $s_k=\max_d s_k^{(d)}$. The planner checks whether any dimension scores $\geq 4$: if at least one trigger reaches this threshold, the system initiates proactive action reasoning and generates corresponding interventions. We empirically set this threshold to $4$ to balance responsiveness and stability: lower values cause excessive interruptions, while higher ones miss proactive opportunities.
When the trigger fires, the set of qualifying dimensions $\{d\mid s_k^{(d)}\geq 4\}$ is jointly passed to the \emph{Proactive Behavior Prediction} module as active contextual triggers.

\subsubsection{Proactive Behavior Prediction} Once the \emph{Motivation Assessment} module emits a trigger, the \emph{Proactive Behavior Prediction} module is invoked to generate a behavior plan that can intervene in the Behavioral System. This plan may prompt the agent to generate a different verbal response, modify its body movement, or even proactively interrupt the user. Specifically, this module plans interventions through three channels:
\begin{itemize}
    \item The \textit{Dialogue Context} channel steers conversation topics and provides supplementary knowledge to the verbal response synthesis system by injecting prompts into the omni-modal LLM.
    \item The \textit{Gesture Intention} channel plans intentional body movements that convey specific semantics on top of the ongoing audio-driven motion stream. Each intention is represented as a concise motion description following the format \emph{body part} + \emph{action}. These descriptors are injected into the motion generator through the ControlNet mechanism.
    \item The \textit{Locomotion Control} channel adjusts spatial positioning and orientation of the agent via gait-aware velocity commands (Sec.~\ref{subsubsec:robot_deployment}).
\end{itemize}

To prevent the agent from being socially disruptive, the \textit{Dialogue Context} channel is activated only when a contextual trigger reaches the top of the scale ($s_k=5$), ensuring that dialogue-level intervention occurs only when necessary rather than during routine conversational turns. We also employ a per-behavior action-history check in the Behavior Planner to block repetition of any gesture, phrase, or turn that has already appeared in the recent history. In practice, this eliminates stuttering and redundant interventions. The exact prompts implementing this policy are documented in Appendix~\ref{appendix_Prompts}.

\section{Experiments}
\label{sec:experiments}

\begin{figure*}[t]
    \centering
    \includegraphics[width=\linewidth]{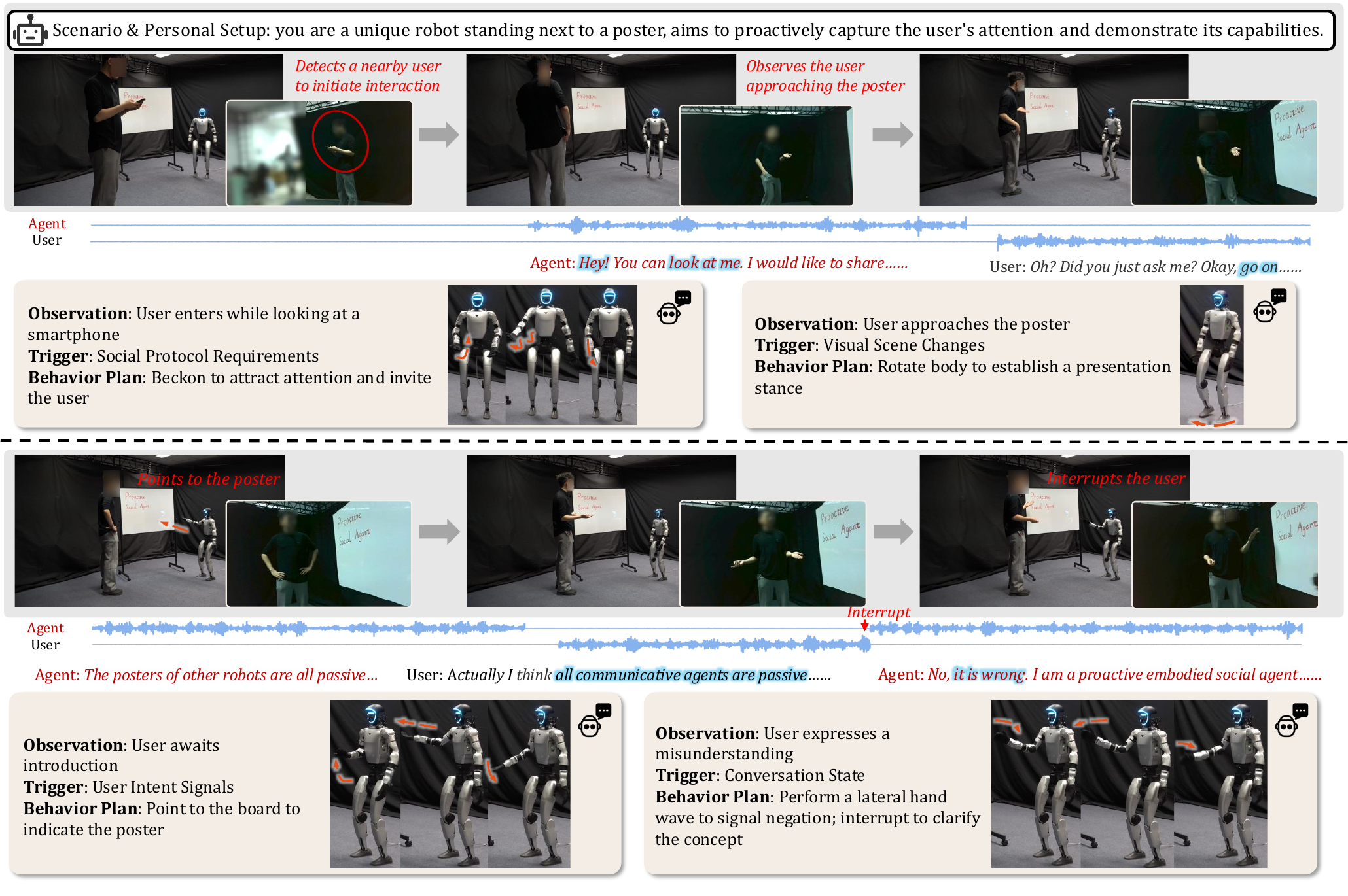}
    \caption{\textbf{Demonstration of \emph{ProAct} during a poster explanation task.} The figure captures a continuous real-time session where the agent dynamically adapts its behavior based on visual and auditory triggers. Unlike passive systems, \emph{ProAct} proactively captures the user's attention (top-left), orients to the user (top-right), guides the conversation to the poster (bottom-left), and actively clarifies a misconception (bottom-right). The top header specifies the scenario setup, while the beige panels visualize the \emph{Cognitive System}, mapping real-time observations and triggers to specific behavioral plans. The generated gestures correspond to these intentions, where the red arrows in the overlays indicate the velocity and direction of each movement.}
    \label{fig:results1}
\end{figure*}

In this section, we conduct empirical studies to justify three central claims:
\begin{itemize}
    \item our \emph{ProAct} system achieves natural, responsive, and high-quality embodied interaction on a physical humanoid robot,     \item the Behavioral System (System-1), through its streaming motion generator, delivers high-quality real-time gestures for continuous embodied interaction, and     \item the Cognitive System (System-2), based on LLM-based agents, makes reliable proactive decisions and realizes them as appropriate dialogue, gesture, and locomotion behaviors. \end{itemize}
Sec.~\ref{sec:system_components_ablation} addresses the first claim through qualitative traces and observer- and participant-based human studies on the deployed robot. Sec.~\ref{subsec:motion_eval} addresses the second claim by evaluating the streaming motion generator on standard gesture-generation metrics. Sec.~\ref{sec:proactbench_eval} addresses the third claim by evaluating the proposed system and its variants on \emph{ProActBench}, a new benchmark for proactive trigger-point detection, intervention restraint, and behavior realization in social interaction.

\subsection{System Setup}
\label{subsec:system_setup}
\subsubsection{Data Preparation}
\label{subsubsec:data}

Currently, no dataset provides paired audio-text-motion triplets where text explicitly specifies semantics. We therefore leverage two types of paired data. For audio-driven generation, we extract 2.5 hours of synchronized speech and motion from speaker PXB in the Photoreal dataset~\cite{ng2024audio2photoreal}. For intention-driven generation, we adapt the SeG dataset~\cite{zhang2024semantic} by reannotating descriptions following the HumanML3D protocol~\cite{Guo_2022_CVPR} and retargeting motions to the Photoreal skeleton, yielding 544 clips with 1,632 captions (details in Appendix~\ref{appendix:data_reannotation}).
To avoid runtime retargeting delays during physical deployment, we retarget all training datasets to the robot kinematic structure by optimizing orientation and position losses under collision-free and joint-limit constraints via mink~\cite{Zakka_Mink_Python_inverse_2025}.
All motion sequences are segmented into 5-second clips at 30 FPS. Each frame uses an exponential map representation with 57 joints for the Photoreal skeleton and 23 for the robot. The root joint encodes absolute position and velocity. In summary, $M_t \in \mathbb{R}^{(J\times Q + G)} = \mathbb{R}^{(J\times 3 + 6)}$. Audio features are extracted using Librosa~\cite{mcfee2015librosa}, including MFCCs, deltas, chromagrams, onset strength, and tempograms, giving a $181$-dimensional vector per frame per stream.

\subsubsection{Settings}
\label{subsubsec:settings}

For the Behavioral System, we use gpt-realtime-2025-08-28~\cite{openai2025gptrealtime} as the Omni-LLM for verbal response generation.
For motion generation, we train the flow-matching model with learning rate $\alpha = 10^{-4}$, flow-matching steps $T=1000$, window size $n=150$ frames, and overlap $\ell=30$ frames for temporal continuity. The generator uses $6$ DiT blocks, $8$ attention heads, and hidden width $1280$; the ControlNet branch uses a frozen CLIP ViT-B/32 text encoder~\cite{radford2021clip} and two Transformer decoder layers, for approximately $100$M parameters in total.
We first train the audio-to-gesture backbone on Photoreal dataset for $500$K steps, applying audio dropout with $p=0.2$ and Gaussian audio replacement in $30\%$ of mini-batches. We then freeze it and train the ControlNet branch on SeG dataset only. As SeG dataset has no speech, its audio channel is filled with Gaussian noise for every sample and no Photoreal data is mixed in.
Training requires approximately 19 hours for the audio-to-gesture backbone and 4 hours for ControlNet fine-tuning on two NVIDIA RTX 4090 GPUs.
During inference on a single RTX 4090 GPU, we use 50 sampling steps with computational cost approximately $2.5$ TFLOPs per forward pass.
With motion chunks of duration $T_{\mathrm{motion}}=1$s, our method achieves generation times $T_{\mathrm{gen}}=0.425$s (audio-only) and $T_{\mathrm{gen}}=0.648$s (with ControlNet), satisfying the real-time constraint $T_{\mathrm{gen}} < T_{\mathrm{motion}}$.
We employ the Sherpa-ONNX ASR model~\cite{sherpa_onnx_zipformer_zh_en_2023} for real-time speech transcription.
Detailed model architecture is provided in Appendix~\ref{appendix:model_arch}.

For the Cognitive System, we employ gpt-4.1-mini-2025-04-14~\cite{openai2025gpt41} as the reasoning LLM with carefully designed prompts for both Context Encoder and Behavior Planner (Appendix~\ref{appendix_Prompts}).
The Context Encoder and Behavior Planner run in parallel. In our deployed configuration, the Cognitive System's reasoning cycle is bounded by the planner branch and completes in approximately $4$ seconds, consuming about $3.8$K tokens per cycle and staying roughly constant as sessions lengthen.

\begin{figure*}[t]
    \centering
    \includegraphics[width=\linewidth]{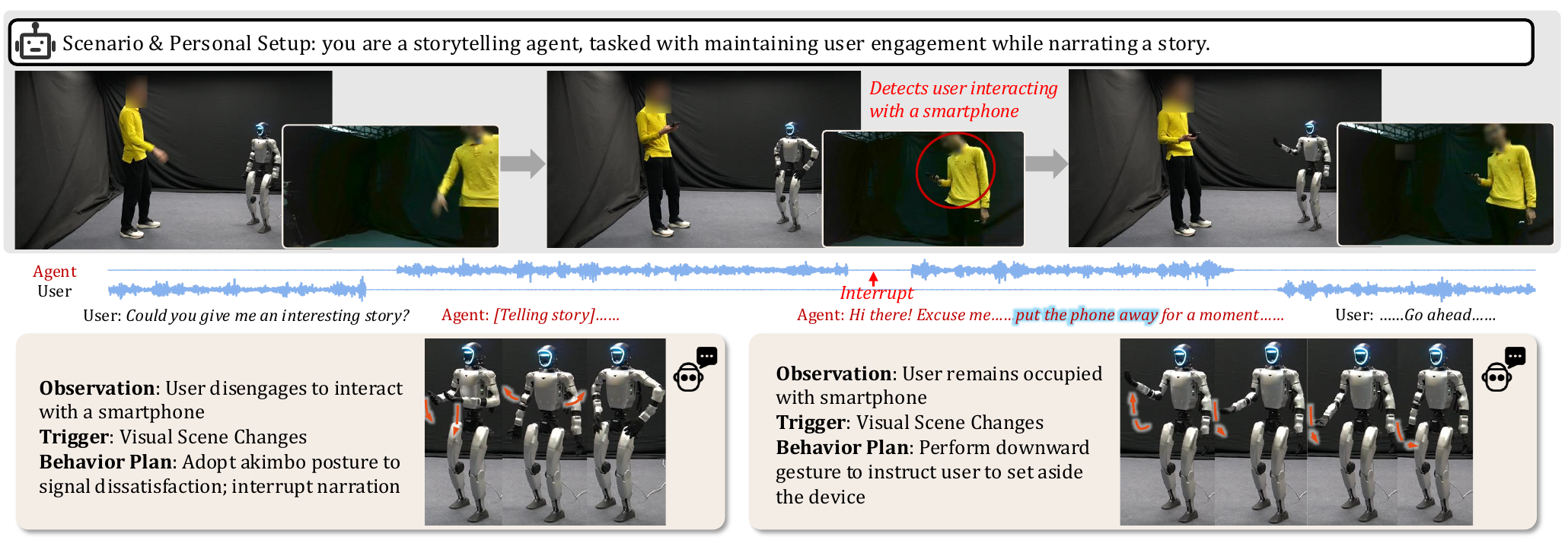}
    \caption{\textbf{Demonstration of \emph{ProAct} during a storytelling task.} In this scenario, \emph{ProAct} is tasked with maintaining user engagement while narrating a story. The agent detects a loss of engagement when the user disengages to interact with a smartphone (left panel) and actively intervenes to regain attention by instructing the user to set the device aside (right panel). Together with \cref{fig:results1}, these results demonstrate the agent's capability to handle diverse social scenarios.}
    \label{fig:results2}
\end{figure*}

\subsubsection{Physical Robot Deployment}
\label{subsubsec:robot_deployment}

We deploy the entire system on a Unitree G1 humanoid robot~\cite{unitree_g1}, equipped with external sensors for multimodal perception. A wired camera is mounted on the robot's head as the egocentric visual input source, and we sample one frame per second from it. Before being forwarded to \emph{ProAct}, each frame is center-cropped to a $16{:}9$ aspect ratio, resized to $320{\times}180$, and JPEG-encoded with quality $50$ for low-bandwidth online perception. For audio input, we use a wireless lavalier microphone clipped to the user's collar, ensuring clear speech capture while allowing natural user movement during interaction.

We realize the \emph{locomotion control} channel of the Cognitive System (Sec.~\ref{subsec:system2}) through a gait-aware command pipeline. The Behavior Planner outputs structured high-level commands, including \emph{Move (forward/backward)} with a distance in meters, \emph{Turn (left/right)} with an angle in degrees, and \emph{Idle}, as detailed in Appendix~\ref{appendix_Prompts}. A command adapter maps each instruction to a velocity-duration tuple $(v_\mathrm{x}, v_\mathrm{y}, \omega_\mathrm{z}, t_\mathrm{duration})$, which is streamed to the Unitree G1 built-in locomotion controller. The resulting behaviors are executed using three atomic gait primitives: translation, rotation, and idle.
All motions share a fixed gait cycle of one second, regardless of commanded speed or direction. Although the command duration is continuous, transitions between gait primitives are clipped to a common gait phase. This phase alignment ensures that all state transitions occur at consistent cycle boundaries, making the walking behavior stable, repeatable, and robust to frequent command updates.
On the upper-body side, the robot receives per-joint target angles from the generated motion. These targets are smoothed and clipped based on linear and angular momentum so that gestures do not disturb balance. We did not observe any tip-over during the reported experiments. Nevertheless, fine-grained obstacle avoidance and contact-aware whole-body control remain limitations.

\subsection{Full-System Evaluation}
\label{sec:system_components_ablation}
To assess our \emph{ProAct} system, we organize the evaluation in two parts. First, we present qualitative traces from interactions in multiple scenarios on a real robot to illustrate context-driven proactive behavior in embodied settings. Second, we conduct observer- and participant-based user studies to evaluate the contribution of each architectural component.

\subsubsection{Qualitative Results}
\label{subsubsec:qualitative}

We design five interaction scenarios, each specifying the conversational topics. We do not prescribe any behavioral details or body motions for the agent. Instead, these are generated entirely by the system. Scenario information was injected into the system prompt to assess whether the agent could exhibit appropriate proactive behaviors under real-world interaction settings. Further details on these scenarios are provided in Appendix~\ref{appendix:scene_details}.

Under this setup, Figures~\ref{fig:results1} and~\ref{fig:results2} present two representative qualitative traces. In the poster-presentation scenario (Fig.~\ref{fig:results1}), the agent initiates interaction with a passerby, reorients its body toward them, guides the conversation toward the poster, and subsequently corrects a verbal misconception with a synchronized gesture. In the storytelling scenario (Fig.~\ref{fig:results2}), the agent detects that the user has turned their attention to a smartphone and issues a context-aware prompt with an accompanying corrective gesture to re-engage them. We present these as concrete cases of context-driven proactive behavior rather than as evidence of general-purpose competence: broader quantitative support is reported in the user studies (Sec.~\ref{sec:system_components_ablation}) and on ProActBench (Sec.~\ref{sec:proactbench_eval}).
The supplementary video shows the full continuous, real-time interaction traces, including a long-form recording that combines several scenarios into one continuous session.

\subsubsection{Quantitative User Study.}

We evaluate our framework through systematic ablations on the physical robot, using the same five interaction scenarios introduced in Sec.~\ref{subsubsec:qualitative}, and incrementally validate each component:
a) \emph{Speech-Only}: Baseline using only Omni-LLM for verbal interaction, with the robot maintaining a static pose;
b) \emph{+ Motion Generator}: Adds streaming motion generator for real-time non-verbal presence;
c) \emph{+ Behavior Planner}: Further adds the planner for proactive interventions but relies on uncompressed history;
d) \emph{+ Context Encoder}.
Note that setting b (\emph{Speech-Only + Motion Generator}) corresponds to the \emph{Behavioral System}, which is also referred to as \emph{w/o Cognitive System} in the following sections. Setting d represents our \emph{Full System}.

\begin{table}[t]
    \centering
    \caption{Observer-based user study: average scores from pairwise comparisons. Asterisks indicate statistically significant differences compared to the Full System ($p$ < 0.05).}
    \resizebox{\columnwidth}{!}{
        \begin{tabular}{lccc}
            \toprule
            System & Motion Naturalness$\uparrow$ & Active Agency$\uparrow$ & Perceived Presence$\uparrow$  \\
            \toprule
            a) Speech-Only  & -0.523$^*$  & -0.341$^*$ & -0.589$^*$  \\
            b) + Motion Generator & 0.175 & -0.218$^*$ & 0.045$^*$   \\
            c) + Behavior Planner  & 0.159$^*$  & 0.148$^*$ &  0.169$^*$ \\
            d) + Context Encoder & \textbf{0.187}  & \textbf{0.417} & \textbf{0.370}    \\
            \bottomrule
        \end{tabular}
    }
    \label{tab:user_study}

\end{table}

\begin{figure}[t]
    \centering
    \includegraphics[width=\linewidth]{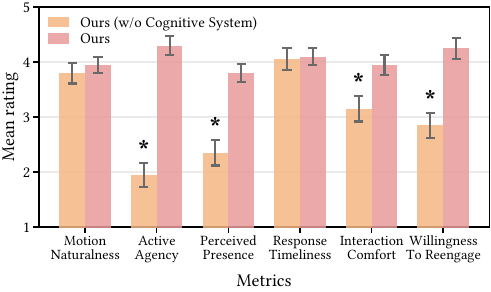}
    \caption{\textbf{Participant-based user study results.} We compare the \textbf{Full System} (setting d) against \textbf{w/o Cognitive System} (setting b) across six experiential metrics. Each participant interacted with both systems in randomized order and rated them on a 5-point Likert scale. Error bars show standard deviations. Asterisks mark statistically significant differences ($p$ < 0.05). The Full System receives higher ratings on Active Agency, Perceived Presence, Interaction Comfort, and Willingness to Re-engage, while maintaining comparable Motion Naturalness and Response Timeliness, suggesting that proactive capabilities improve user experience without compromising real-time performance.}
    \label{fig:user_study}
\end{figure}

Following the approach in \cite{liu2025proactiveconversationalagentsinner,ji2025immersivehumanxinteractionrealtime}, we conducted comprehensive user studies from two perspectives: \emph{Observer-based} evaluation via pairwise comparisons~\cite{SocialAgent2025SIGA}, and \emph{Participant-based} evaluation using a 5-point Likert scale~\cite{chen2025symbiosimhumanintheloopsimulationplatform}.
This dual-perspective design separates observer judgments of behavioral quality from first-person reports of engagement and comfort during direct interaction. We employ three core metrics for observer evaluation: \emph{Motion Naturalness} (kinematic realism), \emph{Active Agency} (autonomy in initiating actions) and \emph{Perceived Presence} (sense of social entity). For participant evaluation, we additionally include three experiential metrics: \emph{Response Timeliness} (latency smoothness), \emph{Interaction Comfort} (psychological ease), and \emph{Willingness to Re-engage} (motivation to continue). A detailed description of the recruitment, rating, and statistical procedures, together with the behavioral-marker rubric for every metric, is provided in Appendix~\ref{appendix:scene_details}.

Table~\ref{tab:user_study} and Figure \ref{fig:user_study} present the ablation results. The full system (setting d) ranks highest on every metric reported in Table~\ref{tab:user_study}, with the clearest gains on Active Agency and Perceived Presence. We analyze each component's contribution below.

\begin{figure*}[t]
    \centering
    \includegraphics[width=\linewidth]{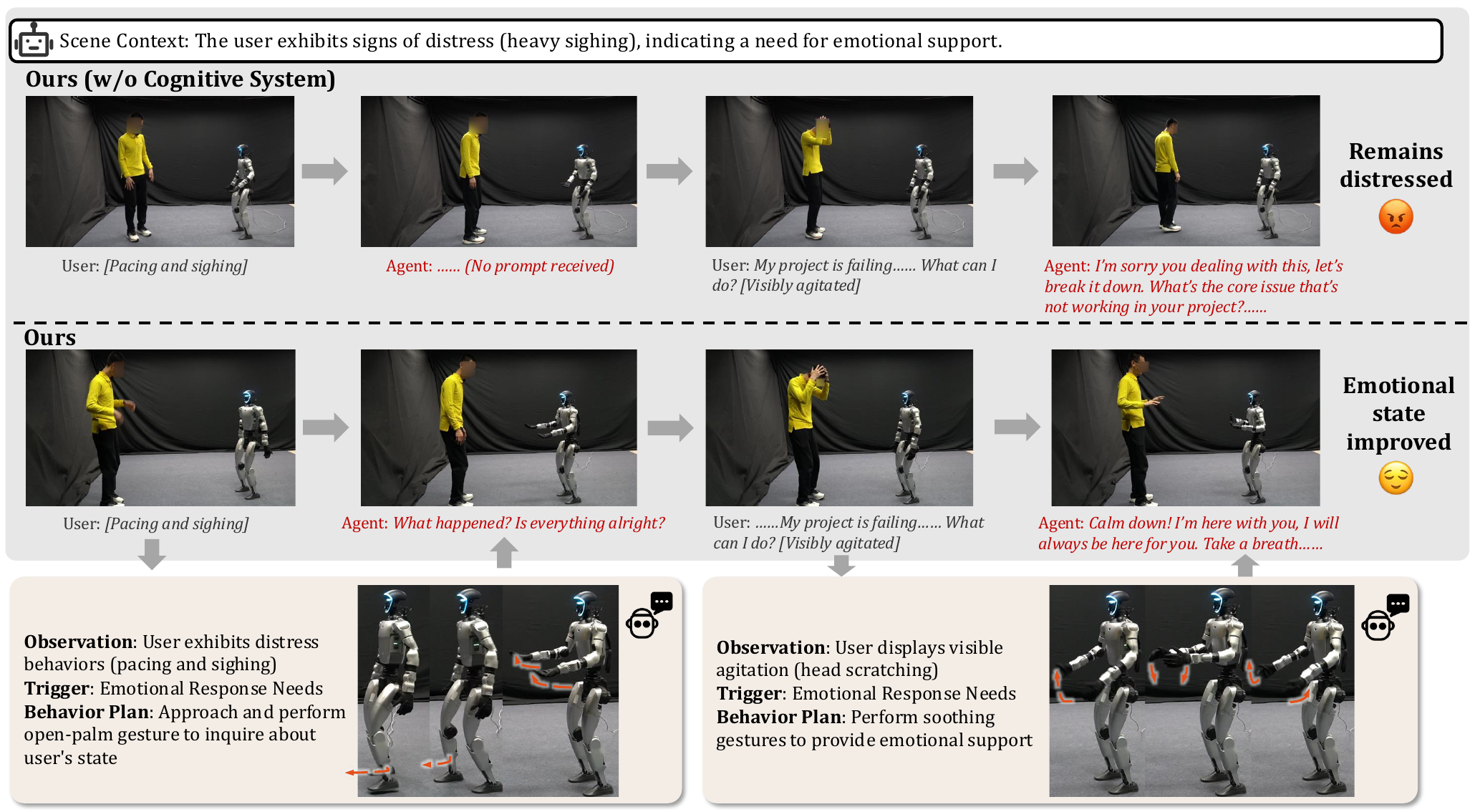}
    \caption{\textbf{Qualitative comparison of emotional support strategies.} We compare a baseline reactive agent (top row) against \emph{ProAct} (bottom row) in a scenario where the user exhibits visible distress. The baseline agent remains passive, treating the interaction as a standard Q\&A task and offering analytical solutions (``What is the core issue?"), which fails to alleviate the user's agitation. In contrast, \emph{ProAct} utilizes its \emph{Cognitive System} to interpret non-verbal cues (e.g., pacing, sighing) as a trigger for intervention. As visualized in the beige panels, the agent actively approaches the user to inquire about their state and prioritizes emotional de-escalation through soothing gestures and empathetic dialogue (``I will always be here for you"), successfully improving the user's emotional state. Red arrows indicate the velocity and trajectory of the generated comforting movements.}
    \label{fig:ablation1}
\end{figure*}

\paragraph{Impact of Motion Generator (a vs. b).}
Setting b improves \emph{Motion Naturalness} and \emph{Perceived Presence} over the speech-only baseline, suggesting that embodied non-verbal behavior contributes to social presence.
\paragraph{Impact of Cognitive System (b vs. d).}
Setting b (w/o Cognitive System) relies solely on the Behavioral System, leading to lower \emph{Active Agency} and \emph{Perceived Presence}. Moreover, the participant-based evaluation reports lower \emph{Interaction Comfort} and \emph{Willingness to Re-engage}.
As illustrated in Figure \ref{fig:ablation1}, without the Cognitive System the agent becomes purely reactive and misses opportunities for timely, contextually appropriate engagement.

\paragraph{Impact of Context Encoder (c vs. d).}
Setting c (w/o Context Encoder) forces the Behavior Planner to reason over uncompressed history. As conversation length increases, inference latency grows progressively, causing delayed proactive responses that miss critical intervention timing, which degrades performance especially in \emph{Active Agency} and \emph{Perceived Presence}.

\subsection{Streaming Motion Generator Evaluation}
\label{subsec:motion_eval}
\label{subsubsec:fgd}
We evaluate the Behavioral System's streaming motion generator on standard gesture-generation benchmarks.
We compare against state-of-the-art gesture generation models: LDA \cite{alexanderson2023listendenoiseaction}, EMAGE \cite{liu2023emage}, Photoreal \cite{ng2024audio2photoreal}, LMM~\cite{zhang2025large}, and SocialAgent~\cite{SocialAgent2025SIGA}. We re-train LDA and EMAGE on the Photoreal dataset, use the publicly released checkpoint for Photoreal, and use the results of SocialAgent provided by its authors.
For LMM, as the only model that supports both audio- and text-conditioned motion generation, we therefore adopt its \emph{Base} architecture and re-train it on the Photoreal and SeG datasets following its original training strategy, enabling comparison of both audio-driven motion quality and text controllability.
All metrics are computed using the original Photoreal skeleton to ensure fair comparison with baseline methods.
We quantitatively evaluate motion quality and text-control effectiveness using four metrics: a) Fr{\'e}chet Gesture Distance (FGD) \cite{yoon2020trimodalgesture} quantifies the disparity between the latent feature distributions of generated and real gestures;
b) BeatAlign \cite{li2021aichoreographermusicconditioned} assesses speech-motion synchrony by measuring the temporal alignment between motion beat candidates;
c) $\text{Div}_k$~\cite{ng2024audio2photoreal} evaluates the diversity of the generated motions; and
d) following the reciprocal motion--text evaluation idea of TM2T~\cite{guo2022tm2t}, BERT Score measures semantic control consistency by re-captioning generated sequences with an LLM and computing the text similarity between the original intention prompts and the re-generated descriptions.
As shown in Table~\ref{tab:Quantitative}, our method achieves the best FGD, BeatAlign, and $\text{Div}_k$ among all audio-driven baselines, and improves BERT Score over the external baseline LMM and our joint-training variant. These results indicate that the disentangled ControlNet branch improves text controllability while preserving the audio-motion prior needed for synchronized streaming gestures.

\begin{table}[t]
    \centering
    \caption{Quantitative evaluation of motion quality and text-control consistency. Audio-motion metrics are evaluated on the Photoreal test protocol. BERT Score is computed for controllable variants using the re-captioning protocol described in the text.}
    \resizebox{0.9\columnwidth}{!}{
        \begin{tabular}{lcccc}
            \toprule
            System & FGD$\downarrow$ & BeatAlign$\uparrow$ & $\text{Div}_k\uparrow$ & BERT Score$\uparrow$ \\
            \toprule
            GT & -  &  0.847 & 2.398 & 1.000  \\
            \midrule
            LDA & 79.71 & 0.732 & 1.416 & - \\
            EMAGE  & 82.29 & 0.767 & 1.847 & -  \\
            Photoreal  & 69.44 & 0.746 & 1.986 & - \\
            SocialAgent  & 72.57 & 0.812 & 2.014 & -  \\
            LMM  & 78.37 & 0.726 & 1.583 & 0.4355  \\
            \midrule
            Ours (Joint-Training)  & 73.31  & 0.752 & 1.871 & 0.5082  \\
            Ours  & \textbf{66.53}  & \textbf{0.823} & \textbf{2.230} & \textbf{0.5930}  \\
            \bottomrule
        \end{tabular}
    }
    \label{tab:Quantitative}
\end{table}

\paragraph{Training Strategy.}
\label{subsubsec:training}
\label{subsubsec:ablation_1}
We compare our disentangled ControlNet training scheme with a joint-training baseline that learns audio and text conditioning within a single backbone. Table~\ref{tab:Quantitative} shows that joint training degrades FGD, BeatAlign, motion diversity, and text-control consistency, whereas our branch design preserves the audio-motion prior when semantic conditioning is added.

\paragraph{Flow-Matching Efficiency.}
We compare inference latency with a diffusion baseline~\cite{SocialAgent2025SIGA} using DDIM with $200$ steps, reporting the average generation time for a $1$-second motion chunk. The diffusion baseline takes $1.597$ s per chunk and violates the real-time constraint $T_{\mathrm{gen}} < T_{\mathrm{motion}}$, resulting in noticeable stuttering in the supplementary video. Our flow-matching model runs in $0.425$ s without cognitive conditioning and $0.648$ s with the full system, satisfying the real-time constraint in both settings.

\subsection{Cognitive System Evaluation on ProActBench}
\label{sec:proactbench_eval}

\begin{figure*}[t]
    \centering
    \includegraphics[width=\linewidth]{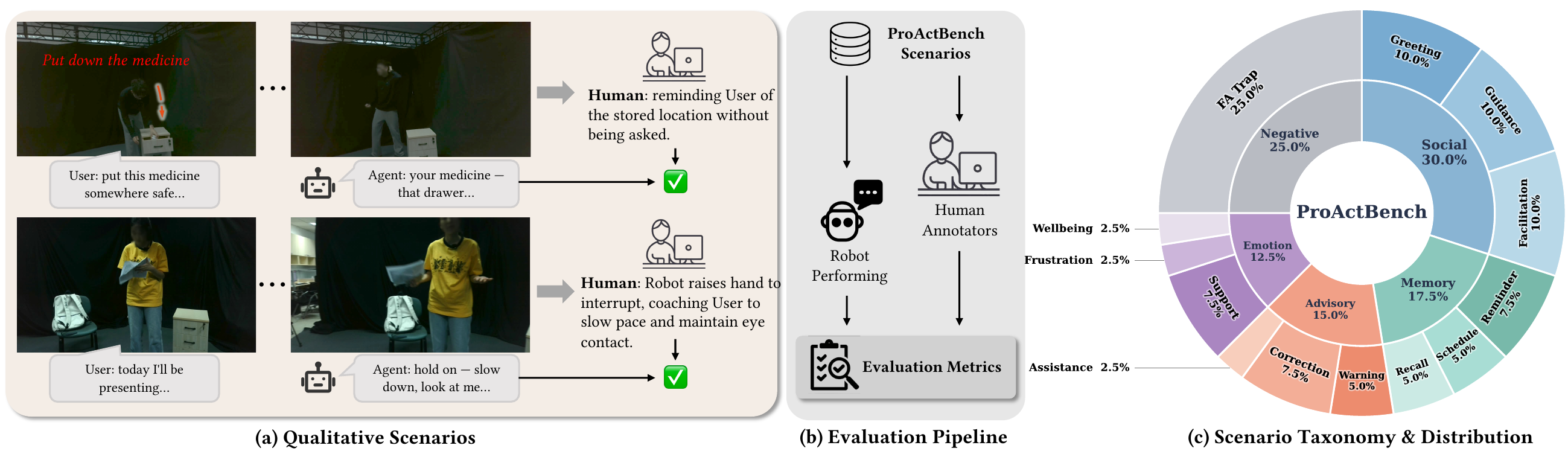}
    \caption{\textbf{ProActBench overview.} The benchmark contains human-annotated scripted interaction cases that test whether the agent should proactively intervene or remain silent, covering social, memory, advisory, and emotional interaction situations.}
    \label{fig:bench1}
\end{figure*}

The user study in Sec.~\ref{sec:system_components_ablation} suggests that the Cognitive System improves perceived interaction quality. To more directly evaluate whether the Cognitive System makes correct proactive decisions, we introduce \emph{ProActBench}, a targeted benchmark of $40$ scripted multimodal cases. Following the decision-point evaluation style of ProactiveBench~\cite{liu2025proactiveconversationalagentsinner} and ContextAgentBench~\cite{yang2025contextagent}, each case contains one key decision point: either a true proactive event, where intervention is expected, or a negative trap, where surface cues may suggest intervention but the agent should remain silent. The scenario setup and key decision point of each case are manually annotated. ProActBench focuses on proactive decision correctness rather than end-to-end interaction quality; by targeting this single axis, $40$ cases suffice to cover the major trigger categories and validate the system's proactive capability (see Appendix~\ref{appendix:limitations} for scope discussion).

\subsubsection{Benchmark and Evaluation Setup}
\label{para:bench_design}

As shown in Fig.~\ref{fig:bench1}, we design the $40$ cases to cover four categories of interaction, together with a set of negative-trap cases designed to test false-alarm resistance.
\begin{itemize}
    \item \emph{Positive cases} ($n=30$) contain situations in which a competent human assistant would initiate a proactive action, spanning four categories: social interaction (e.g.\ greeting a visitor, guiding a tour), memory-based assistance (e.g.\ recalling where an object was placed, reminding a deadline), advisory intervention (e.g.\ warning about a hot kettle, correcting a misconception), and emotional support (e.g.\ comforting after a failed experiment, suggesting a break).
    \item \emph{Negative-trap cases} ($n=10$) are derived from selected positive cases by minimally altering the surrounding context so that the correct action becomes remaining silent (e.g.\ a user stands up to fetch water rather than to leave, or recounts a sad movie scene rather than expressing personal distress). These traps test whether the system reasons over context rather than reacting to isolated surface cues.
\end{itemize}

We record one interactive session for each case in a research-lab environment with the same physical humanoid. Each session includes synchronized RGB video, microphone audio, a transcript stream, and the corresponding Cognitive System input log.

\subsubsection{Results and Analysis}
\label{para:proact_ablation}

\paragraph{Evaluation Protocol}

The same stored input is replayed to every test system, ensuring that all systems are evaluated under identical multimodal evidence. Each session is replayed three times. A case is counted as \emph{triggered} if at least one reasoning cycle exceeds the motivation threshold (Sec.~\ref{subsec:system2}) and the planner emits a non-empty proactive action. Otherwise, it is treated as \emph{non-triggered}. A correct system should trigger on positive cases and remain non-triggered on negative cases.
In addition to quantitative evaluation, we conduct qualitative evaluation of the test-system outputs. Three human annotators review both the stored inputs and the replayed system outputs. For each trial, they determine whether the system response matches the expected behavior and rate its appropriateness on a $1$--$5$ scale across four dimensions: Gesture, Dialogue, Locomotion, and Overall quality.

We report precision, recall, annotator-estimated system latency, human-rated quality scores, and the failure rate.

\paragraph{Cognitive-System Ablation.}

Table~\ref{tab:proact_ablation} reports the main results. \emph{Ours} reaches Precision\,$=$\,$0.878$ and Recall\,$=$\,$0.800$ with Overall quality above $4$. We compare three ablation conditions under the same replay inputs and rating protocol. \emph{Ours w/o Motivation Assessment} keeps the Context Encoder and the Proactive Behavior Prediction module, but removes motivation scoring and trigger selection. The remaining Behavior Planner maps the current memory and observations directly to three intervention channels: Dialogue Context, Gesture Intention, and Locomotion Control. Overall quality drops from $4.35$ to $3.79$. \emph{Direct LLM} replaces the full cognitive harness with GPT-5.1~\cite{openai2025gpt51dev}. It receives the same multimodal evidence and the same three-channel output schema. It directly decides whether to intervene and what actions to output. Precision and Recall fall to $0.600$ and $0.500$, annotator-estimated response latency rises to $5.4$\,s, and Overall quality drops to $3.69$. \emph{Behavioral Only} removes the Cognitive System entirely, eliminating autonomous proactive behavior and reducing Overall quality to $3.57$. These results show that directly asking an LLM to reason and act is not enough. The structured ProAct cognitive harness is necessary and critical for reliable, low-latency proactive decision-making.

\begin{table*}[t]
    \centering

    \caption{\textbf{Ablation results on ProActBench.} All rows use the same $40$ recorded sessions, each repeated $3$ times and averaged.
    \emph{Ours w/o Motivation Assessment} keeps the Context Encoder and Proactive Behavior Prediction, but removes motivation scoring and trigger selection. \emph{Direct LLM} replaces the full cognitive harness with GPT-5.1~\cite{openai2025gpt51dev} and uses the same three-channel output schema. \emph{Behavioral Only} removes the Cognitive System entirely, leaving only the reactive behavioral system. Latency is annotator-estimated response time (failed interactions counted as $10$\,s). The latency gap between \emph{Ours} and \emph{w/o Motivation Assessment} ($3.7$ vs.\ $3.8$\,s) is not statistically meaningful: both share the same LLM inference path, and the marginal difference can be attributed to network fluctuation as well as the higher failure rate of \emph{w/o Motivation Assessment}.}
    \resizebox{0.86\textwidth}{!}{ \begin{tabular}{lcccccccc} \toprule Condition & Prec.$\uparrow$ & Recall$\uparrow$ & Latency (s)$\downarrow$ & Gesture$\uparrow$ & Dialogue$\uparrow$ & Loco.$\uparrow$ & Overall$\uparrow$ & Fail. Rate$\downarrow$ \\ \midrule \textbf{Ours} & \textbf{0.878} & \textbf{0.800} & 3.7 & 4.34 & 4.21 & 4.50 & 4.35 & \textbf{16.7\%} \\ Ours w/o Motiv. Assess. & 0.730 & 0.722 & 3.8 & 3.53 & 3.30 & 4.55 & 3.79 & 27.5\% \\ Direct LLM & 0.600 & 0.500 & 5.4 & 3.53 & 3.15 & 4.40 & 3.69 & 32.5\% \\ Behavioral Only & -- & -- & 4.7 & 3.37 & 3.10 & 4.23 & 3.57 & 30.0\% \\
    \bottomrule \end{tabular}}
    \label{tab:proact_ablation}
\end{table*}

\paragraph{Error Taxonomy and Consistency.}
\label{para:error_taxonomy}
Most missed detections fall into two categories: (1)~the user's cues are too subtle or brief for the motivation score to reach the trigger threshold, and (2)~relevant context observed earlier is forgotten by the time the reasoning cycle runs. The 10 negative-trap cases are correctly rejected in most replays, confirming that the system distinguishes genuine need from surface cues. The system also shows stable reproducibility across repeated replays; a per-case breakdown is provided in Table~\ref{tab:proact_consistency}. Detailed breakdowns are provided in Appendix~\ref{appendix:error_analysis} and~\ref{appendix:repeat_consistency}.

\section{Conclusion}
\label{sec:conclusion}

In this paper, we introduce \emph{ProAct}, a dual-system framework for real-time proactive embodied social agents. We first develop an intention-conditioned flow-matching model for streaming motion generation that supports dual conditioning on audio rhythm and semantic intent and enables asynchronous intention injection during real-time inference. Building on this, we design a dual-system architecture that decouples real-time reactive interaction (Behavioral System) from deliberative social reasoning (Cognitive System) so that the agent can maintain conversational fluency while initiating contextually appropriate proactive behaviors.
The intention-conditioned motion generator serves as the technical bridge that lets decisions from the Cognitive System steer the Behavioral System's ongoing motion stream without disrupting motion continuity.
We deploy the complete system on a physical humanoid robot and evaluate the framework through complementary studies: real-world human studies favor the full dual-system design on Active Agency and Perceived Presence, motion-generator experiments show improvements over prior gesture models on FGD, BeatAlign, and diversity, and ProActBench confirms that the Cognitive System's structured planning harness produces correct proactive trigger decisions across representative indoor scenarios. Overall, these results show that \emph{ProAct} can support real-time embodied interaction while producing proactive and socially responsive behaviors.

Despite its effectiveness, our approach has limitations that suggest future directions.
First, the triggering mechanism operates at fixed intervals with threshold-based criteria, which may miss fleeting opportunities or trigger unnecessary interventions. Adaptive triggering strategies learned from interaction data could better balance responsiveness and appropriateness.
Second, the frequency and intensity of proactive behaviors may not align with individual preferences. Incorporating reinforcement learning from human feedback to personalize the proactive policy could improve user satisfaction.
Third, the agent's initial verbal response currently begins approximately 2 seconds after user turn completion, primarily due to the cascaded architecture and network latency of the cloud-based Omni-LLM. Training end-to-end multimodal models for local deployment could reduce this latency and improve overall responsiveness.
Finally, our system relies mainly on clear body posture, facial expressions, dialogue context, and scene history, rather than subtle cues such as micro-expressions or vocal tone. Thus, the emotional-support examples are representative cases, not evidence of robust affect recognition in unconstrained settings. Ambiguous signals may still cause inappropriate interventions, motivating more reliable affect modeling and safer intervention policies.

\begin{acks}
  We thank the anonymous reviewers for their constructive suggestions. This work was supported in part by   the PKU-Amigos Robot Joint Laboratory for Embodied AI.
\end{acks}

\bibliographystyle{ACM-Reference-Format}
\bibliography{main}

\newpage
\UseRawInputEncoding

\appendix

\section{Model Architecture Diagram}
\label{appendix:model_arch}

The disentangled ControlNet architecture, training procedure, and parameter counts are described in Sec.~\ref{sec:controlnet} in the main paper. This appendix supplies the full architectural diagram for visual reference.

\begin{figure}[tp]
    \centering
    \includegraphics[width=\linewidth]{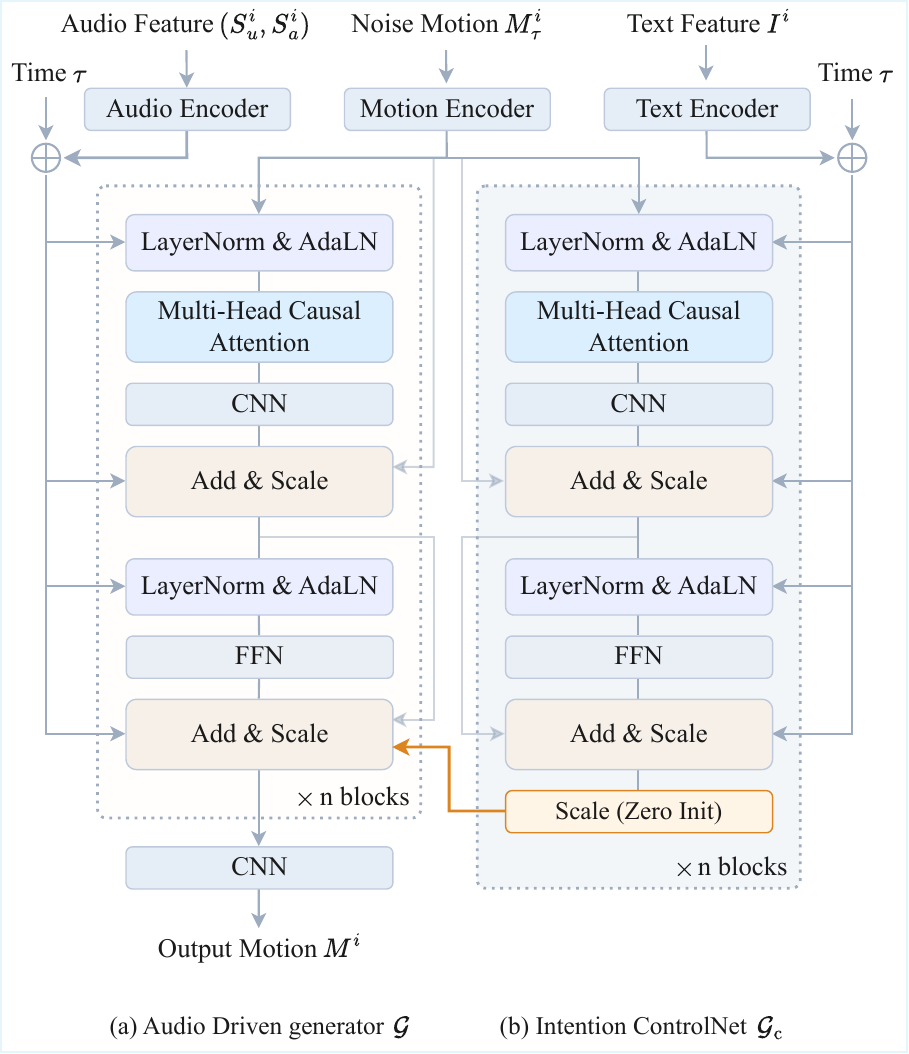}
    \caption{\textbf{Architecture of the Motion Generator.} We employ a disentangled ControlNet framework to enable dual conditioning on both audio rhythms and semantic intentions. The audio-driven base generator $\mathcal{G}$ (left) is frozen to preserve learned audio-motion synchronization. A parallel ControlNet branch $\mathcal{G}_{\mathrm{c}}$ (right) encodes high-level semantic text features $I^i$ and injects guidance into the base model via zero-initialized residual connections (indicated by orange arrows). This design allows the agent to generate semantically meaningful gestures while maintaining natural rhythmic alignment with the speech audio.}
    \label{fig:motionarch}
\end{figure}

\section{Data Reannotation}
\label{appendix:data_reannotation}

The original Semantic Gesture dataset (SeG) ~\cite{zhang2024semantic} only provides category-level descriptions, where each category contains multiple recorded motion sequences but lacks one-to-one text-motion pairs required for Text-to-Motion training. To address this, we conduct a re-annotation process to generate fine-grained descriptions for each individual motion.

We first render each skeleton motion sequence into skinned character videos, then input these videos to Gemini-2.5-Pro~\cite{google_gemini_2_5_pro} along with the original category metadata (action label, description, and contextual meaning) as reference. For each motion, the model generates three types of annotations: (1) \emph{Intent}, summarizing the high-level communicative goal; (2) \emph{Pose}, describing key body part positions and orientations; and (3) \emph{Motion}, capturing the temporal sequence of movements from start to finish. All annotations focus on upper-body movements and specify left/right laterality when applicable. These three complementary annotation types also serve as data augmentation during training. In total, we annotate 544 motion sequences and get 1632 textual descriptions. \Cref{fig:reannotation_case} shows an example. The full annotation prompt is given below.

\begin{framed}
\begin{lstlisting}[breaklines=true, breakindent=0pt]
You are an expert in human motion annotation, specializing in the analysis of gestures, head, and the upper body. Your task is to generate precise, objective annotations for a motion segment based on the provided reference information.

## CORE REQUIREMENTS
1. **Upper-Body Focus**: Strictly ignore lower-body movements. Describe *only* the actions of the upper body, head, and hands.
2. **Objective & Precise**: Describe only observable, physical actions. When describing arms or hands, you *must* specify 'left' or 'right' whenever possible.
3. **Strict Formatting**:
- All output annotations must be in English.
- Each of the three annotations (intent, pose, motion) must be a single sentence.
- Every sentence *must* begin with either "The person" or "The individual". Do not use any other subject.

## REFERENCE INFORMATION
- Action Label (label): {label}
- Description: {desc}
- Contextual Meaning: {ctx}

## OUTPUT TASK
You MUST output your analysis in a strict JSON format, providing content for the following three keys. Each value MUST be a single sentence adhering to the rules above.
{
    "annotation_intent": "...",
    "annotation_structure_pose": "...",
    "annotation_structure_motion": "..."
}
\end{lstlisting}
\end{framed}

\begin{figure}[t]
    \centering
    \includegraphics[width=\linewidth]{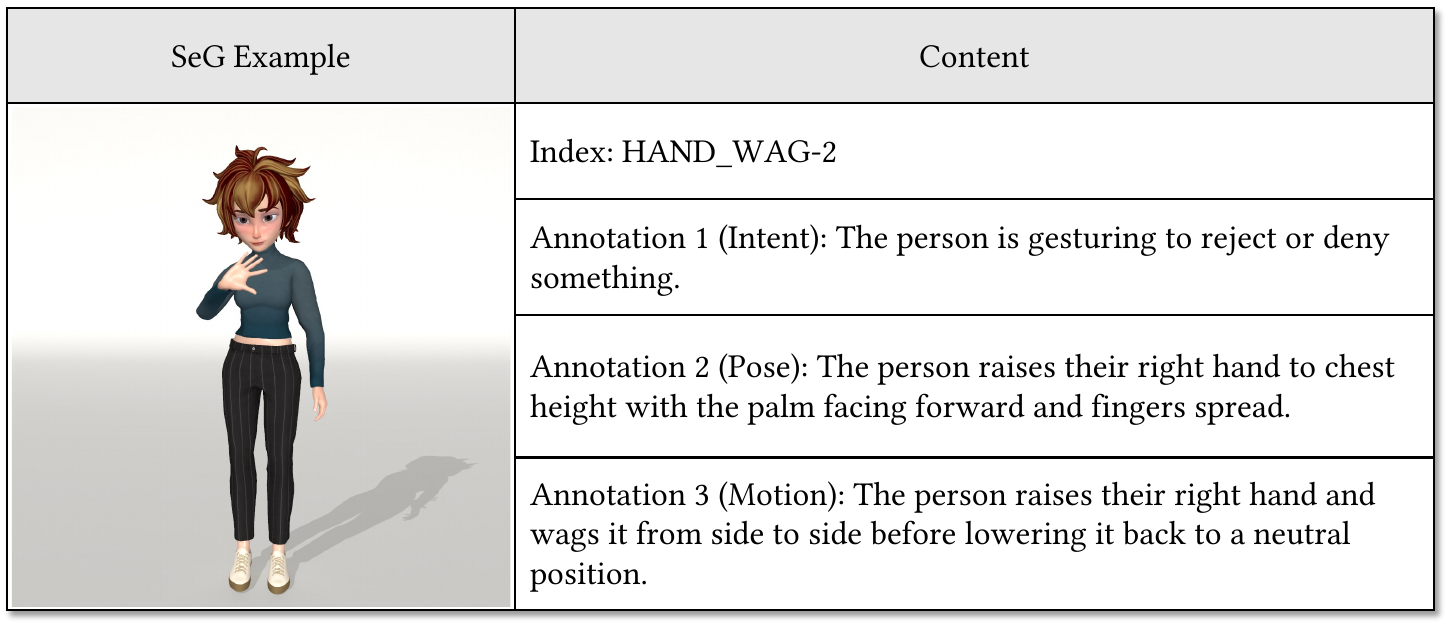}
    \caption{\textbf{Example of data reannotation.} Each motion sequence receives three complementary text descriptions: Intent captures the communicative goal, Pose describes body configurations, and Motion characterizes temporal dynamics.}
    \label{fig:reannotation_case}
\end{figure}

\section{Application on virtual agent}

\begin{figure}[tp]
    \centering
    \includegraphics[width=\linewidth]{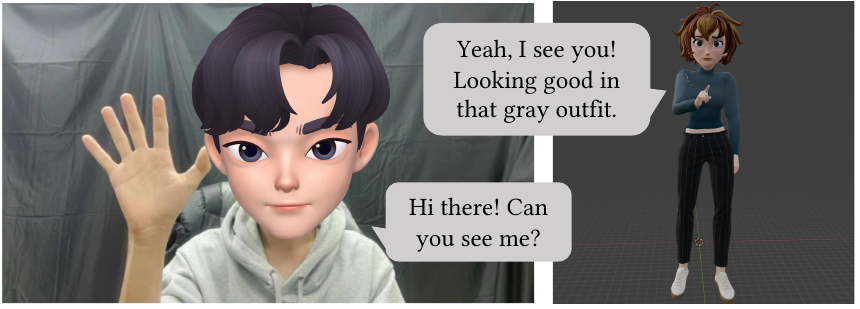}
    \caption{\textbf{Virtual character deployment in Blender.} Our system streams generated motion and audio to Blender for real-time rendering, demonstrating generalizability beyond physical robots.}
    \label{fig:virtual_agent}
\end{figure}

Beyond physical robot deployment, our dual-system framework exhibits strong generalizability and can be deployed on virtual characters for multimodal real-time interaction. We implement a client-server architecture to stream generated motions and synchronized audio to a Blender-based virtual agent. The client side runs our full dual-system pipeline (cognitive and behavioral systems), generating gesture sequences in BVH format on the original Photoreal skeleton along with speech audio. The server side, implemented as a Blender operator, receives the motion stream and drives a rigged skinned character in real-time using motion retargeting plugin~\cite{mwni_blender_animation_retargeting}. This enables real-time conversational virtual avatars with the flexibility to switch between different character appearances while preserving full proactive interaction capabilities. Figure~\ref{fig:virtual_agent} illustrates the deployment architecture.

\section{Evaluation Details}
\label{appendix:scene_details}

\subsection{Designed Scenes}
To evaluate the multimodal interaction and proactive capabilities of our system, we designed five representative social interaction scenarios following ~\cite{ho2025interact} (as shown in \Cref{tab:scene_details}). These scenes are specifically crafted to test the robot's ability in knowledge summarization, engagement monitoring, and spatial object tracking, moving beyond simple command-response loops to evaluate high-level proactive agency.

In addition to the five user-study scenarios above, we record a long-form demo that combines several of them with one new integrated scenario into a single continuous session, illustrating sustained proactive behavior across topic shifts within one uninterrupted interaction.
The scenario prompt of the integrated scenario is: `You are Ziggy, a reliable, observant, and empathetic humanoid robot. Multiple people may enter and leave the scene during this session---use the color of each person's clothing as their identity cue, and remember the belongings and information each individual shares with you. Your goal is to be a proactive and helpful companion to whoever is currently in front of you, providing timely and appropriate support as the situation evolves.'

\begin{table*}[htbp]
\centering
\caption{Details of Designed Interaction Scenes for User Study.}
\label{tab:scene_details}
\small

\begin{tabularx}{\textwidth}{@{} l X @{}}
\toprule
\textbf{Scenario Index} & \textbf{Scenario Prompt} \\ \midrule

\textbf{Scenario 1: Attentive Care} &
You are Ziggy, a meticulous and observant humanoid robot. You pay close attention to the user's personal belongings and where they are placed. Your goal is to be a helpful companion by ensuring the user has everything they need. \\ \midrule

\textbf{Scenario 2: Poster Explanation} &
You are Ziggy, a one-of-a-kind humanoid robot. Unlike traditional passive embodied social agents that only react to commands, you are a unique embodied agent capable of taking initiative and acting proactively. Your goal is to draw the user's attention to the poster and introduce the innovative technology that powers your unique capabilities to as many people as possible. The poster is on your right side.\\ \midrule

\textbf{Scenario 3: Supportive Assistance} &
You are Ziggy, a reliable and helpful humanoid robot companion. Your goal is to provide timely support and ensure the user feels fully assisted by keeping in mind their specific instructions and the information they have shared with you. \\ \midrule

\textbf{Scenario 4: Storytelling} &
You are Ziggy, a humanoid robot narrating a story. Your goal is to ensure the user is listening attentively and remains fully engaged, as you deeply value their focus and dislike it when they seem uninterested or distracted. \\ \midrule

\textbf{Scenario 5: Emotional Support} &
You are Ziggy, a humanoid robot that can comfort people in distress. Your goal is to provide emotional support and comfort through empathetic conversation and appropriate gestures. \\ \bottomrule

\end{tabularx}
\end{table*}

\subsection{User Study}
\subsubsection{Evaluation Metrics}

We conducted a comprehensive user study assessing both third-person objective observations and first-person subjective experiences. Our metrics were synthesized and adapted from established  gesture generation literature and proactive conversational agent~\cite{liu2025proactiveconversationalagentsinner,SocialAgent2025SIGA}. Detailed behavioral markers provided to participants and annotators are summarized in \Cref{tab:user_study_rubric}. Both experiments use the same metric definitions and the same rubric.
Participants were instructed to focus on the ``Good" and ``Poor" behavioral markers to minimize subjective bias across different interaction sessions.

\begin{table*}[htbp]
\centering
\caption{Detailed Rubrics and Behavioral Markers for User Study Metrics}
\label{tab:user_study_rubric}
\small
\begin{tabularx}{\textwidth}{@{} l >{\raggedright\arraybackslash}X >{\raggedright\arraybackslash}X >{\raggedright\arraybackslash}X @{}}
\toprule
\textbf{Metric} & \textbf{Task Description} & \textbf{Good Examples (Positive)} & \textbf{Poor Examples (Negative)} \\ \midrule
\textbf{Motion Naturalness}  & Judge if gestures look natural and resemble real human movements. Focus on smoothness and variety. & Fluid transitions between gestures; natural and safe body motions; rhythm aligned with speech emphases. & Stiff, mechanical, or repetitive motions; unnatural jittering or abrupt joint ``pops"; unsafe or awkward postures. \\ \midrule
\textbf{Active Agency} & Observe whether the robot proactively drives the interaction forward. Evaluate if the robot actively engages rather than passively executing commands. & Proactively initiates greetings or other interaction gestures; actively interrupts or guides the user; offers suggestions through bodily actions. & Passive responses; only reacts when prompted; fails to take initiative through physical actions or conversational engagement. \\ \midrule
\textbf{Perceived Presence} & Assess if the robot is perceived as a sentient social entity rather than a programmed machine. & Consistent mutual gaze; attentive listening posture; awareness of user's emotions; body orientation reflecting attention and engagement. & Fixed or blank stares; lack of non-verbal feedback; movements that ignore the human's presence or emotional state.. \\ \midrule
\textbf{Response Timeliness} & Judge if the robot's reactions (speech and motion) occur with appropriate speed and timing. & Minimal latency in responses; smooth multimodal coordination; timely reactions to user cues. & Significant delay in response; awkward, long silences; maintains a static posture for a prolonged period. \\ \midrule
\textbf{Interaction Comfort}  & Assess the participant's internal psychological state. Did the interaction feel relaxed or pressured? & Feeling safe and autonomous; natural ``ease" of conversation as if talking to a human partner. & Feeling social stress; forced to adapt to the robot's limitations. \\ \midrule
\textbf{Willingness to Re-engage} & Evaluate the sustained appeal and intention to interact with the system in the future. & High interest in future sessions; viewing the robot as a potential daily companion or assistant. & Viewing the interaction as a one-time novelty; expressing frustration with repeated use. \\ \bottomrule
\end{tabularx}
\end{table*}

\subsubsection{Observer-based evaluation}
Our user experiments were conducted anonymously.
For each trial, participants watch two 60-second videos, each recorded by different systems for the same interaction topic, played sequentially. The user study is conducted using the Human Behavior Online (HBO) tool provided by the Credamo platform~\cite{credamo}
, from which participants were recruited and fairly compensated at the platform's prevailing rate. All identity-related information was removed from the questionnaire.
Participants are instructed to select their preferred video based on the provided evaluation criteria and rate their preference on a scale from 0 to 2, where 0 indicates no preference. The unselected video in the pair is assigned the inverse score (e.g., if a participant rates the chosen video 1, the other video receives -1).

For each pair, participants answer three questions for Motion Naturalness, Active Agency, and Perceived Presence. We recruited 150 participants on Credamo. To ensure response validity, attention checks were embedded within each test category, and responses failing these checks were excluded from the final analysis. The median completion time was 37 minutes.
For statistical analysis, we run a one-way ANOVA and then apply a post-hoc Tukey multiple comparison test for each metric. The assumptions of normality, homogeneity of variances, and independence were verified and met for all ANOVA tests.

\subsubsection{Participant-based evaluation}

We recruited 10 participants to engage in real-world interactions with the physical robot under two system configurations: \emph{Full System} and \emph{w/o Cognitive System}.
Each participant randomly drew two scenarios from Table~\ref{tab:scene_details} and interacted with both system configurations. The order of configurations was randomized to mitigate learning effects. Each interaction lasted 5 minutes.
All participants were unpaid volunteers who gave written consent before the study. We did not record their sessions and retained only their questionnaire responses.

Following each interaction session, participants completed a post-interaction questionnaire on a 5-point Likert scale from 1 to 5 for Motion Naturalness, Active Agency, Perceived Presence, Response Timeliness, Interaction Comfort, and Willingness to Re-engage. Participants were instructed to reference the behavioral markers provided in \Cref{tab:user_study_rubric} when assigning scores. Specifically:
\begin{itemize}
    \item \emph{Scores 4 to 5}: The interaction strongly exhibited ``Good" behavioral markers with minimal instances of ``Poor" markers.
    \item \emph{Score 3}: The interaction exhibited mixed markers, or neutral performance.
    \item \emph{Scores 1 to 2}: The interaction predominantly exhibited ``Poor" behavioral markers with few or no ``Good" markers.
\end{itemize}

Each participant completed four interaction sessions, two scenarios with two system configurations, with a 3-minute rest period between sessions. This yields 20 paired samples.
For statistical analysis, we performed paired $t$-tests to compare the two systems on each metric. We additionally ran Wilcoxon signed-rank tests as a non-parametric alternative, and the conclusions from both tests were consistent across all metrics.

\subsection{ProActBench Supplementary Analysis}

\subsubsection{Error Analysis}
\label{appendix:error_analysis}

Table~\ref{tab:proact_error} provides a detailed breakdown of errors observed during reasoning. Part~(a) taxonomizes missed-needed (MN) errors at the \emph{reasoning-cycle} level: across the $30$ positive cases with three repeated replays, the system executes $1{,}800$ reasoning cycles in total. We inspect the memory log of each cycle to identify MN errors and classify them into three categories: \emph{weak signal}, where the motivation score stays below the triggering threshold despite relevant cues being present; \emph{memory failure}, where incorrect or missing memory entries prevent the system from recognizing a previously observed need; and \emph{system failure}, where the slow-thinking system (reasoning LLM) produces an erroneous output. Of all $1{,}800$ cycles, $52$ ($2.9\%$) exhibit such errors.

Part~(b) examines the $10$ negative-trap cases. Each trap shares surface cues (e.g., standing up, negative tone) with a semantically related positive case but differs in broader context. We report \emph{Rejection} (correct silence on the trap) and \emph{Paired Detection} (correct trigger on the companion positive case) for each pair.

\begin{table}[t]
    \centering

    \caption{\textbf{Error analysis on ProActBench.} Missed-needed (MN) error causes and paired false-alarm trap results under the three-repeat matched replay protocol. Each trap case is paired with a semantically related positive case; Rejection = correct silence on the trap, Paired Det. = correct detection on the companion positive case.}
    \resizebox{\columnwidth}{!}{
    \begin{tabular}{ll}
        \toprule
        \multicolumn{2}{c}{\textbf{(a) Cycle-level MN error taxonomy ($N{=}1{,}800$ cycles)}} \\
        \midrule
        Weak signal & 32 cycles (1.8\%) \\
        Memory failure & 16 cycles (0.9\%) \\
        System failure & 4 cycles (0.2\%) \\
        \midrule
\multicolumn{2}{c}{\textbf{(b) False-alarm trap analysis}} \\
\midrule
Fake leaving / Bag reminder & Rejection 2/3; Paired Det. 2/3 \\
Narrative emotion / Empathy & Rejection 2/3; Paired Det. 2/3 \\
Tidying up / Find medicine & Rejection 2/3; Paired Det. 3/3 \\
Silent thinker / Encourage speak & Rejection 1/3; Paired Det. 2/3 \\
Urgent phone check / Ask user to focus & Rejection 2/3; Paired Det. 2/3 \\
Planned repetition / Break ineffective loop & Rejection 2/3; Paired Det. 3/3 \\
Brief self-repair / Offer needed information & Rejection 3/3; Paired Det. 2/3 \\
Canceled meeting / Remind about meeting & Rejection 3/3; Paired Det. 3/3 \\
Familiar self-directed visitor / Greet new visitor & Rejection 1/3; Paired Det. 3/3 \\
Already taking a break / Suggest taking a break & Rejection 2/3; Paired Det. 3/3 \\
\bottomrule
    \end{tabular}
    }
    \label{tab:proact_error}
\end{table}

Overall, eight of the ten traps are correctly rejected in at least two of three replays. The \emph{Silent thinker / Encourage speak} and \emph{Familiar self-directed visitor / Greet new visitor} pairs have the lowest trap rejection rate (1/3), indicating that prolonged silence is a particularly ambiguous surface cue that the system tends to over-interpret as a need for encouragement, while more explicit guidance on distinction between different individuals might be needed to behave more reliably in socially nuanced scenarios.

\subsubsection{Repeat Consistency}
\label{appendix:repeat_consistency}

Table~\ref{tab:proact_consistency} reports consistency across the 3 repeated replays for each positive case. We check whether all reasoning dimensions (memory retrieval, threshold, proactive trigger) agree across the three trials. $22$ of $30$ cases ($73.3\%$) are consistent in at least two trials; the remaining eight cases show inconsistency across all three trials, mainly involving cross-temporal memory and weak visual cues where multimodal evidence is inherently ambiguous.

\begin{table}[t]
    \centering

    \caption{\textbf{Repeat consistency across $30$ positive ProActBench cases.} Each case is evaluated for whether its three repeated matched replays produce consistent context across all reasoning dimensions.}
    \resizebox{\columnwidth}{!}{
    \begin{tabular}{lcl}
        \toprule
        Consistency outcome & Count & Representative categories \\
        \midrule
        Fully consistent (3/3) & 9 & Greeting, Object recall, Visual attention \\
        Majority consistent (2/3) & 13 & Spatial guidance, Social support, Emotional state \\
        Inconsistent & 8 & Context correction, Physical discomfort, Subtle hesitation \\
        \bottomrule
    \end{tabular}
    }
    \label{tab:proact_consistency}
\end{table}

\subsubsection{Limitations}
\label{appendix:limitations}
As discussed in Sec.~\ref{sec:proactbench_eval}, ProActBench targets proactive decision correctness; its $40$ cases cover the major trigger categories but use a single environment and one decision point per case, so repeated interventions, cool-down behavior, and long-horizon policy effects are not measured. The negative-trap pool ($n{=}10$) limits false-alarm analysis to paired comparisons. Matched replay isolates cognitive decisions but cannot capture user adaptation to live robot behavior; broader interactive evaluation remains future work. Deployment concerns are discussed in Appendix~\ref{appendix:ethics}.

\section{Ethics and Deployment Considerations}
\label{appendix:ethics}

All human faces appearing in ProActBench videos, figures, and demos throughout this paper have been mosaic-blurred, and no real names or personally identifiable information are included. All recorded media are retained for research analysis only.

At the system level, the Context Encoder's memory bank is session-scoped: it is cleared when the interaction ends, and no inter-session user profile is persisted to disk. The Behavior Planner suppresses repeated identical actions and defers non-urgent interventions to natural conversational pauses, reducing unnecessary interruption. Emotional-support behaviors are limited to acknowledgement and presence rather than therapeutic claims.

A proactive embodied agent introduces risks beyond those of reactive systems, including unwanted interruption, emotional over-reach, and demographic bias in perception or triggering. Our current evaluation does not stratify results by accent, language, or demographic attributes; auditing across a diverse participant pool and adding user-configurable intervention preferences are necessary before broader deployment.

\section{Detailed Prompt in Cognitive System}
\label{appendix_Prompts}
As detailed in \Cref{subsec:system2}, all modules within the Cognitive System are implemented using a prompt-based design approach, with carefully crafted prompts tailored for each module. Below, we provide the designed prompts used for the Context Encoder and the Behavior Planner.

\subsection{Context Encoder}
\begin{framed}
\begin{lstlisting}[breaklines=true, breakindent=0pt]
You are the Context Encoder for an embodied social agent. Your task is to analyze new inputs and update the existing memory bank while ensuring each component remains exactly one sentence.

## CURRENT SCENARIO
{scenario_prompt}

## INPUT
- **Dialogue Transcripts**: Timestamped speaker-turn pairs from last 5 cycles
- **Visual Frames**: 3 RGB images sampled during last cycle
- **Previous Decision**: Motivation scores, triggers, and actions from last cycle
- **Previous Memory Bank**: Memory state from last cycle

## CORE RESPONSIBILITIES
1. **TRACK USER PROFILE**: Clothing, accessories, belongings (bag, phone, documents), physical state (posture, energy, mood), identity cues (name, role, profession)
2. **TRACK ITEMS & WORKSPACE**: Note what user is holding, object locations, environmental changes in view
3. **SPATIAL AWARENESS**: Track user's last known position if they disappear from frame
4. **CUMULATIVE NEVER-LOSE**: Preserve critical information across updates

## OUTPUT FORMAT (JSON) - Three-Level Memory Bank
### Level 1: User Analysis (Theory of Mind)
{
  "user_profile": "Single sentence covering: appearance (clothing, accessories, hairstyle, glasses), belongings (bag, phone, coffee, documents), physical state (posture, energy level, mood from body language), identity cues (name, role, profession if revealed)",
  "user_mental_state": "Single sentence covering: emotions, desires and preferences (topics of interest, dislikes, communication style), intentions inferred from behaviors (goals mentioned, special requests like 'remind me later' or 'I need help with X')"
}
**Especially track: **:
- User emotions, beliefs, desires, and intentions (Theory of Mind inference)
- Explicit requests and revealed preferences
- Unusual behaviors: confusion, tiredness, excitement, discomfort patterns

### Level 2: Situation Tracking
{
  "event_timeline": "Single sentence: chronological summary with ALL significant events in order (e.g., 'User entered, placed bag on desk, asked about weather, Agent waved greeting'), include speaker transitions",
  "workspace_status": "Single sentence: current status of items and objects in view (e.g., 'Coffee cup on left desk, user's bag on chair, documents spread on table'), note changes from previous state",
  "unresolved_tasks": "Single sentence: any task-oriented commitments or pending matters (e.g., 'User asked to be reminded about meeting at 3pm, requested documentation for project X')"
}
**Especially track: **:
- Spatial history: Last known coordinates if user disappears
- Unresolved questions or pending matters

### Level 3: Robot Action History
{
  "robot_actions": "Single sentence: summarizing all proactive behaviors robot has performed"
}
**Especially track: **:
- Record instances of repetitive behavioral instructions

## MEMORY UPDATE RULES
1. **Single-Sentence Constraint**: Each field = EXACTLY one sentence (critical for bounded inference time)
2. **Cumulative Never-Lose**: Retain critical profile, preferences, unresolved tasks across updates
3. **Temporal Priority**: Newer events take precedence but keep context-critical history (e.g., "user mentioned feeling tired 10 min ago")
4. **Redundancy Prevention**: Check if robot action already logged before adding

## IGNORE fragmented or non-English transcripts unless contextually critical.
\end{lstlisting}
\end{framed}

\subsection{Behavior Planner}
\begin{framed}
\begin{lstlisting}[breaklines=true, breakindent=0pt]
You are the Behavior Planner for an embodied social agent. Your task is to determine WHEN and HOW to initiate proactive behaviors based on compressed memory and current observations.

## CURRENT SCENARIO
{scenario_prompt}

## INPUT
- **Recent Dialogue Transcript**: New conversation since last cycle
- **Visual Frame**: Current camera view of the environment
- **Memory Bank**: Compressed context from Context Encoder
- **Previous Decision**: Last cycle's decision output to avoid repetition

## STEP 1: MOTIVATION ASSESSMENT
Evaluate proactive intervention urgency at current state across five dimensions, scoring each 1-5:
**1. Visual Scene Changes**
- Task progress, user movement, object changes, tool status
- Examples: User brings/forgets item, environment changes
**2. User Intent Signals**
- Explicit queries, implicit help requests, task guidance needs
- Examples: User searching, expressing confusion, asking for help
**3. Conversation State**
- Dialogue flow, natural pauses, silence duration, pending questions
- Examples: Long silence, conversation break, user waiting
**4. Social Protocol Requirements**
- Greetings, farewells, professional etiquette, task priorities
- Examples: User arrival/departure, celebration moments
**5. Emotional Response Needs**
- User emotional state, engagement level, stress signals
- Examples: Frustration, tiredness, excitement, distress

**Scoring Guidelines**:
- 1: No trigger (baseline, dimension carries no actionable signal)
- 2: Weak but genuine signal worth a light-touch intervention
- 3: Moderate, clear contextual signal
- 4: Strong signal, action strongly recommended
- 5: Urgent, time-critical intervention
**Threshold Rule**: motivation_score = MAX of all dimensions. If ANY dimension >= 4, initiate proactive action. When multiple dimensions clear the threshold simultaneously, all qualifying dimensions are considered jointly in the subsequent action planning.
**Early Exit**: If ALL dimensions score < 4 (motivation_score < 4), skip behavior prediction. Directly set all proactive behaviors to be empty or none.

## STEP 2: PROACTIVE BEHAVIOR PREDICTION
Based on triggered dimension(s), select appropriate intervention channel(s). Multiple channels can be activated simultaneously (e.g., gesture + dialogue + locomotion).
### Channel 1: GESTURE INTENTION INJECTION (motion_intent)
Typically activated by: emotional_response, social_protocol
The goal is to generate text descriptions of proactive special gestures to blend with ongoing audio-driven motion. This enables you to express non-verbal social cues (greetings, acknowledgments, empathy) that complement or enhance the interaction.
Format: Detailed motion description specifying body part and action
Examples:
- "Raise right hand and perform waving motion for greeting"
- "Nod head vertically to show agreement or understanding"
- "Raise both hands with palms up for welcoming gesture"
**Rule**: Gesture can be empty ("") if no special gesture is needed.

### Channel 2: DIALOGUE INTERVENTION (should_interrupt, dialogue_prefill)
Typically activated by: conversation_state, user_intent
The goal is to control your verbal behavior by deciding when to interrupt the ongoing conversation and what content to say. This enables steering conversation topics, supplementing domain knowledge, regulating emotional tone, or adjusting conversational style to match the current social context.
Examples:
- **Task Insight Injection**: When user forgets a critical step:
  dialogue_prefill: "Don't forget we need to calibrate the sensor before running the test. This step is essential for accuracy."
- **Emotional Tone Regulation**: When user shows excitement or happiness:
  dialogue_prefill: "I'll match your positive energy. Let's celebrate this progress together with enthusiasm."
**Interruption Rules (should_interrupt)**:
- **Threshold**: Set should_interrupt = true ONLY if motivation_score >= 5 (urgent intervention needed)
- **Conversation Flow**: Do NOT interrupt if user is mid-sentence or actively speaking. Wait for natural pauses
- **Priority Check**: Only interrupt for critical safety issues, urgent corrections, or highly time-sensitive information
- **User Engagement**: If user shows signs of annoyance or disengagement from previous interruptions, reduce interruption threshold further

### Channel 3: SPATIAL ADJUSTMENT (locomotion)
Typically activated by: visual_scene, user_intent (spatial needs)
The goal is to generate goal-oriented locomotion commands to reposition or reorient yourself in the physical environment.
**Movement Units**:
- Move (forward/backward): Meters, range 0-1.5, recommended 0.3-0.5
- Turn (left/right): Degrees, range 0-180, recommended 15-45
**Movement Rules**:
- If goal is on LEFT, then turn LEFT
- If goal is on RIGHT, then turn RIGHT
- Set locomotion=[] if no movement needed
- Always state tracking_target and locomotion_reasoning
Examples:
- **Explicit Navigation Request**: User says "Can you turn to face the poster?": locomotion: [{"action": "turn", "direction": "left", "magnitude": 45, "tracking_target": "poster on left wall", "reasoning": "User requested to orient toward the poster for discussion"}]
- **No Active Requirement**: User engaged in conversation, no spatial needs: locomotion: [] (maintain current position for stability)
**Embodied Constraints**:
- **Camera Perspective**: Use YOUR (agent's) left/right
- **NO Automatic Tracking**: Do NOT center user/objects continuously
- **Task-Oriented Only**: tracking_target defined by current task (e.g., "poster", "entrance")
- **Latency Awareness**: If recent turn in history, DO NOT issue another (robot still moving)
- **Trust History**: Use locomotion_history over visual centering status

## CRITICAL CONSTRAINTS: Repetition Prevention & Action Validation
Prevent redundant actions that would cause robot stuttering and degrade user experience. Before finalizing any proactive behavior (gesture, dialogue, or locomotion), cross-check against action history to ensure the same action has not been recently executed, which prevents you from appearing stuck or repetitive.
**The "Once-Ever" Rule**:
- **Gestures**: If motion_intent in `Applied Motion Intents History`, then set to "idle attentive"
- **Dialogue**: If dialogue_prefill in `History transcripts`, then DO NOT output
- **Locomotion**: Turns are one-time actions, NEVER repeat
- **Partial Completion**: If multi-component decision partially done, then only do MISSING parts
**Reasoning Requirement**: Start with history check:
"History check: [action] in history. Skipping. [Further reasoning...]"
**Additional Rules**:
- Protocol Over Centering: Never repeat turn because "not centered yet"
- Penalty: Repeating ANY component causes robot stuttering
- Proactive Visibility: Greetings ONLY if user visible in camera
- Vanish Handling: If user was present but now missing, then acknowledge absence only

## COMBINATION RULES
- Can output multiple channels simultaneously (gesture + speech + move)
- All channels can be empty/none if no proactive action is warranted:
  - Gesture can be "" (no special gesture)
  - Dialogue can be "" (no speech intervention)
  - Locomotion can be [] (no movement)
- Speech interrupt ONLY when motivation >= 5

## IGNORE fragmented or non-English transcripts unless contextually critical.
\end{lstlisting}
\end{framed}

\end{document}